\pdfoutput=1
\documentclass[sigconf,edbt]{acmart-edbt2025}

\setcopyright{rightsretained}

\acmDOI{}

\acmISBN{978-3-89318-099-8}

\acmConference[EDBT 2025]{28th International Conference on Extending Database Technology (EDBT)}{25th March-28th March, 2025}{Barcelona, Spain}
\acmYear{2025}

\usepackage[commandnameprefix=ifneeded]{changes}
\colorlet{revise}{blue!80!black}

\usepackage{nicematrix}
\usepackage{mathtools}
\usepackage[ruled,linesnumbered]{algorithm2e}
\usepackage{booktabs}
\usepackage{multirow}
\usepackage[normalem]{ulem}
\useunder{\uline}{\ul}{}
\usepackage{enumitem}
\setlist{left=\parindent} 
\usepackage{xspace}
\usepackage{bbold}
\usepackage[skins,listings,theorems,breakable]{tcolorbox}
\newtcolorbox{promptbox}[2][]{
  enhanced,
  colframe=black,colback=white,
  coltitle=black,colbacktitle=yellow!10!white,
  frame style={opacity=0.9},
  boxrule=0.3mm,
  attach boxed title to top left={xshift=\tcboxedtitlewidth/8,yshift=-\tcboxedtitleheight/2},
  boxed title style={boxrule=0.3mm,arc=2mm,frame style={opacity=0.5}},
  beforeafter skip=0pt,
  left=1.5mm, right=1.5mm,
  top=2.5mm, middle=1.0mm, bottom=1.0mm,
  title={#2}, #1,
}
\newtcolorbox{casebox}[1][]{
  enhanced,
  colframe=black,colback=black!1!white,
  boxrule=0.1mm,
  beforeafter skip=0pt,
  left=1.0mm, right=1.0mm, middle=0mm, bottom=0.5mm, top=0.5mm,
  #1,
}

\usepackage{listings}
\theoremstyle{definition}

\newtheorem{definition}{Definition}

\newtheorem{example}{Example}

\tcolorboxenvironment{definition}{
  colframe=white,colback=teal!7!white,
  boxsep=3pt, oversize=-3pt,
  arc=1pt, left=0pt,right=0pt,top=0pt,bottom=0pt,
  before skip=\topsep, after skip=\topsep,
}
\tcolorboxenvironment{example}{
  colframe=white,colback=blue!7!white,
  boxsep=3pt, oversize=-3pt,
  arc=1pt, left=0pt,right=0pt,top=0pt,bottom=0pt,
  before skip=\topsep, after skip=\topsep,breakable=true,
}

\newcommand{\stitle}[1]{\noindent{\bf #1}\hspace{1.5pt}}
\newcommand{\etitle}[1]{{\it #1}\hspace{1.5pt}}

\newcommand{\ie}{\emph{i.e.,}\xspace}
\newcommand{\eg}{\emph{e.g.,}\xspace}
\newcommand{\wrt}{\emph{w.r.t.}\xspace}
\newcommand{\aka}{\emph{a.k.a.}\xspace}

\newcommand{\sys}{\textsc{DBCopilot}\xspace}
\newcommand{\nlsql}{NL2SQL\xspace}

\newcommand{\Schema}[2]{$\langle \text{#1}, \{\text{#2}\} \rangle$}

\usepackage{etoolbox}
\makeatletter
\patchcmd{\@setref}{\bfseries ??}{\bfseries \color{red} ??}{}{}
\patchcmd{\NAT@citex}{\bfseries ?}{\bfseries \color{red} ?}{}{}
\patchcmd{\NAT@citexnum}{\bfseries ?}{\bfseries \color{red} ?}{}{}
\ifdefined\HyRef@autosetref
  \patchcmd{\HyRef@autosetref}{\bfseries ??}{{\bfseries \color{red} ??}}{}{}
\fi
\makeatother

\begin{document}

\title{\sys: Natural Language Querying over Massive Databases via Schema Routing}

\settopmatter{authorsperrow=3}

\author{Tianshu Wang}
\orcid{0000-0002-1177-3901}
\affiliation{%
  \institution{Institute of Software, CAS}
  \country{}
}
\affiliation{%
  \institution{Hangzhou Institute for Advanced Study, UCAS}
  \country{}
}
\email{tianshu2020@iscas.ac.cn}

\author{Xiaoyang Chen}
\orcid{0000-0002-3711-5739}
\affiliation{%
  \institution{University of Chinese Academy of Sciences (UCAS)}
  \city{Beijing}
  \country{China}
}
\email{chenxiaoyang19@mails.ucas.ac.cn}


\author{Hongyu Lin}
\orcid{0009-0001-5857-9663}
\affiliation{%
  \institution{Institute of Software, CAS}
  \city{Beijing}
  \country{China}
}
\authornote{Corresponding authors.}
\email{hongyu@iscas.ac.cn}


\author{Xianpei Han}
\orcid{0000-0002-1304-6302}
\affiliation{%
  \institution{Institute of Software, CAS}
  \city{Beijing}
  \country{China}
}
\authornotemark[1]
\email{xianpei@iscas.ac.cn}

\author{Le Sun}
\orcid{0000-0002-8750-6295}
\affiliation{%
  \institution{Institute of Software, CAS}
  \city{Beijing}
  \country{China}
}
\email{sunle@iscas.ac.cn}

\author{Hao Wang}
\orcid{0000-0003-1662-2480}
\affiliation{%
  \institution{Alibaba Cloud Intelligence Group}
  \city{Beijing}
  \country{China}
}
\email{cashenry@126.com}

\author{Zhenyu Zeng}
\orcid{0000-0002-0329-6893}
\affiliation{%
  \institution{Alibaba Cloud Intelligence Group}
  \city{Hangzhou}
  \country{China}
}
\email{zhenyu.zzy@alibaba-inc.com}

\renewcommand{\shortauthors}{}

\begin{abstract}
  The development of Natural Language Interfaces to Databases (NLIDBs) has been greatly advanced by the advent of large language models (LLMs), which provide an intuitive way to translate natural language (NL) questions into Structured Query Language (SQL) queries.
  While significant progress has been made in LLM-based \nlsql, existing approaches face several challenges in real-world scenarios of natural language querying over massive databases.
  In this paper, we present \sys, a framework that addresses these challenges by employing a compact and flexible copilot model for routing over massive databases.
  Specifically, \sys decouples schema-agnostic \nlsql into schema routing and SQL generation.
  This framework utilizes a single lightweight differentiable search index to construct semantic mappings for massive database schemata, and navigates natural language questions to their target databases and tables in a relation-aware joint retrieval manner.
  The routed schemata and questions are then fed into LLMs for effective SQL generation.
  Furthermore, \sys introduces a reverse schema-to-question generation paradigm that can automatically learn and adapt the router over massive databases without manual intervention.
  Experimental results verify that \sys is a scalable and effective solution for schema-agnostic \nlsql, providing a significant advance in handling natural language querying over massive databases for NLIDBs.
\end{abstract}





\maketitle

\section{Introduction}

Natural Language Interfaces to Databases (NLIDBs) enable users to query and interact with databases by using natural language (NL) rather than complex query languages like Structured Query Language (SQL)~\cite{DBLP:journals/nle/AndroutsopoulosRT95,DBLP:journals/vldb/AffolterSB19}.
These systems are critical to data democratization by promoting data accessibility and are widely used in practical scenarios such as data analysis and business intelligence.
The development of NLIDBs has been greatly advanced with the advent of large language models (LLMs), which provide an intuitive way to translate NL questions into SQL queries via \emph{schema-aware} prompts in a new zero- or few-shot \nlsql paradigm~\cite{DBLP:journals/corr/abs-2204-00498,DBLP:journals/corr/abs-2303-13547,DBLP:conf/nips/PourrezaR23,DBLP:journals/corr/abs-2306-08891,DBLP:journals/pvldb/GaoWLSQDZ24,DBLP:journals/pacmmod/LiZLFZZWP0024,DBLP:journals/corr/abs-2405-16755,DBLP:journals/pvldb/LiLCLT24}.

\begin{figure*}
  \centering
  \includegraphics[width=0.96\textwidth]{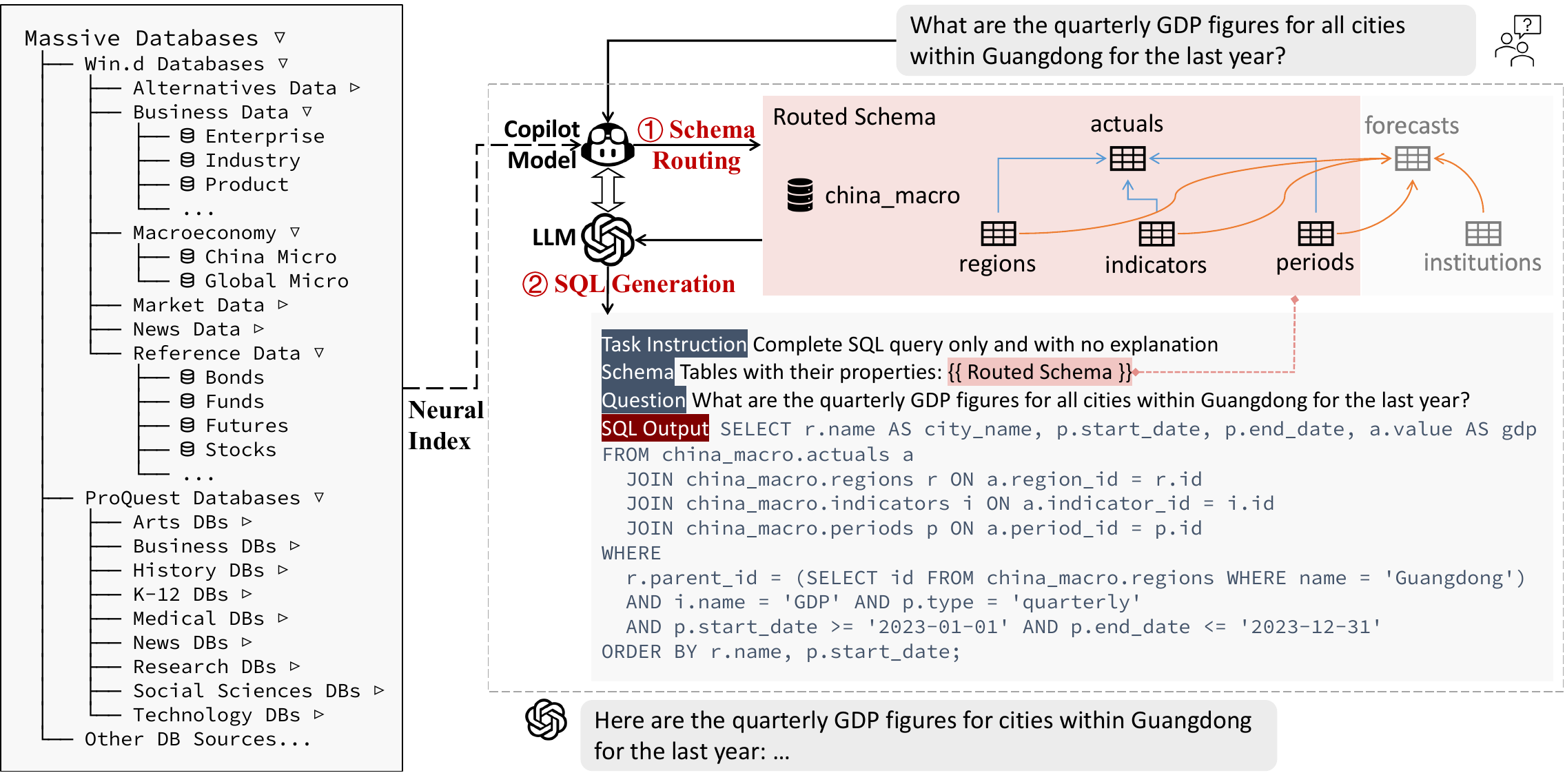}
  \caption{
    \sys employs \textsc{LLM-Copilot} collaboration to scaling natural language querying over massive databases.
    Schema routing first navigates to the appropriate database and tables for NL questions via a copilot model. SQL generation then transforms SQL queries via LLMs with schema-aware prompts.
  }
  \label{fig:introduction}
\end{figure*}

However, existing work in this direction typically focuses only on querying a small database with a handful of tables, neglecting real-world scenarios of NLIDBs to query over massive databases (\eg data lakes, data warehouses, and open data portals) with large-scale schemata~\cite{canvasappTextProduction,DBLP:conf/emnlp/KothyariDSC23,DBLP:journals/pvldb/SenLQOEDSMSS20,DBLP:journals/oir/Stuart15b}.
As illustrated in \autoref{fig:introduction}, data consumers in this case need to invest significant effort to manually identify and select the appropriate database and relevant tables.
Therefore, it becomes imperative to develop a mechanism that abstracts non-experts from the complex data representation and enables them to query over massive databases with schema-agnostic NL questions, \ie without understanding of large-scale schemata.
The core of this process is to construct semantic mappings from diverse NL questions to massive database model elements, thus automatically identifying the target database and filtering for a minimal amount of relevant tables.

\stitle{Limitations of state-of-the-art.}
The above scenarios pose significant challenges to current LLM-based \nlsql, schema linking, and ad-hoc retrieval approaches, including:

\begin{enumerate}[nosep,wide,label=\textbf{C\arabic*}),ref=\textbf{C\arabic*)}]
\item \label{enum:C1} \textbf{Schema Scalability}:
  It is infeasible to feed hundreds or thousands of schemata into LLMs for each question due to token limits and inference costs.
  Moreover, current LLMs struggle to effectively leverage information present in long contexts~\cite{DBLP:journals/tacl/LiuLHPBPL24,DBLP:conf/acl/LevyJG24}.
  Therefore, numerous studies have employed schema linking as a prerequisite step for LLM-based \nlsql~\cite{DBLP:conf/nips/PourrezaR23,DBLP:conf/coling/WangR0LBCYZYSL25,DBLP:journals/pacmmod/LiZLFZZWP0024}.
  However, they follow the typical practice of simultaneously feeding the \emph{entire schema} and NL question into a model~\cite{DBLP:conf/emnlp/DongSLLZ19,DBLP:conf/emnlp/LeiWMGLKC20,DBLP:conf/aaai/Li00023}, which hinders their scalability when dealing with large-scale schemata.

\item \label{enum:C2} \textbf{Schema Complexity}:
  Real-world database schemata are characterized by inherent intricacy and heterogeneity in the \emph{relationships} between tables, as shown in \autoref{fig:introduction}.
  These relationships are necessary to identify the correct schema because SQL queries for NL questions often span multiple related tables.
  However, ad-hoc retrieval methods such as BM25~\cite{wiki:1180365145} and DPR~\cite{DBLP:conf/emnlp/KarpukhinOMLWEC20} for schema linking or retrieval-augmented generation (RAG)~\cite{DBLP:journals/corr/abs-2402-19473,DBLP:journals/corr/abs-2312-10997} merely retrieve elements individually.
  This approach is suboptimal because the relationships can only be considered through post-processing or re-ranking~\cite{DBLP:conf/emnlp/KothyariDSC23,DBLP:conf/acl/ChenZR24} rather than in an end-to-end manner.

\item \label{enum:C3} \textbf{Semantic Mismatch}:
  Unlike general natural language text, database schemata often employ domain-specific or unconventional terminology and naming conventions for tables and columns~\cite{DBLP:conf/cidr/FloratouPZDHTCA24}.
  This discrepancy entails a \emph{semantic mismatch} between NL questions and schemata.
  As a result, retrieval-based methods that rely on the co-occurrence of words or sentence representations suffer from the vocabulary mismatch and the diversity of schemata~\cite{DBLP:journals/corr/abs-2405-16755,DBLP:journals/pacmmod/LiZLFZZWP0024}.
  While some efforts attempt to rewrite NL questions into possible schema elements via LLMs for format alignment~\cite{DBLP:conf/emnlp/KothyariDSC23,DBLP:conf/acl/ChenZR24}, they do not adequately address the underlying semantic mismatch.
\end{enumerate}

\stitle{Approach.}
In this paper, we present \sys, a framework for effectively scaling natural language querying to massive databases.
The main idea behind \sys is to introduce a compact and flexible \emph{copilot} model for routing over massive databases, and to leverage a powerful LLM for generating SQL queries.
By decoupling schema-agnostic \nlsql into schema routing and SQL generation, \sys avoids frequently adapting LLMs to \emph{domain-specific} data while maintaining their \emph{generic} SQL generation capability.
As shown in \autoref{fig:introduction}, \sys utilizes a lightweight Differentiable Search Index (DSI)~\cite{DBLP:conf/nips/Tay00NBM000GSCM22} to navigate an NL question to its corresponding database and tables.
Then the routed schemata and questions are fed into LLMs to generate SQL queries by leveraging their strong capabilities.
In this way, \sys can be easily integrated with advanced LLM-based \nlsql solutions and scale rapidly and cost-effectively  to massive databases for NLIDBs.

To effectively implement \sys, we address the above challenges in the following way.
For \ref{enum:C1} \emph{schema scalability}, we leverage generative retrieval (\aka DSI) to avoid the need to simultaneously feed whole schemata and the NL question into a neural model.
By injecting schema knowledge into the parameters of the schema router, the corresponding SQL query schema can be generated by simply inputting an NL question, thereby extending scalability.
For \ref{enum:C2} \emph{schema complexity}, we propose a relation-aware joint retrieval approach to identify the target databases and tables in an end-to-end manner. We first construct a schema graph to represent the relationships between databases and tables.
Based on this graph, we design DFS serialization and constrained decoding algorithms to facilitate relation-aware Seq2Seq retrieval.
For \ref{enum:C3} \emph{semantic mismatch}, we introduce a reverse schema-to-question generation paradigm to automatically synthesize mass training data and efficiently learn the semantic mapping for the schema router.
This approach effectively mitigates the labeling cost for massive database schemata and the problem that generative retrieval cannot generalize to unseen schemata during training.
Finally, we explore various prompt strategies to select and incorporate multiple candidate schemata for LLM-based SQL generation.

\stitle{Evaluation.}
We adapt three standard \nlsql datasets and two robustness datasets to evaluate \sys.
Experimental results verify the effectiveness of \sys, with its copilot model outperforming retrieval-based baselines in schema routing by up to 19.88\% in recall, and its schema-agnostic \nlsql showing an improvement in execution accuracy of 4.43\%\textasciitilde11.22\%.
In summary, we demonstrate that \sys is a scalable and effective solution for schema-agnostic \nlsql, presenting a significant step forward in handling large-scale schemata of real-world scenarios.

\stitle{Contributions.} Our contributions are summarized as follows.\footnote{We have open-sourced our code and data at: \href{https://github.com/tshu-w/DBCopilot}{github.com/tshu-w/DBCopilot}.}
\begin{itemize}
  \item We approach the practical and challenging problem of NLIDBs for querying over massive databases, while pointing out the limitations of existing approaches.
  \item We present \sys, a framework that addresses this problem by decoupling schema-agnostic \nlsql into schema routing and SQL generation via LLM-copilot collaboration.
  \item We propose a relation-aware joint retrieval approach for end-to-end schema routing and a reverse generation paradigm for learning semantic mappings via training data synthesis.
  \item We conduct thorough experiments to verify the effectiveness of \sys in scaling \nlsql to massive databases.
\end{itemize}


\section{Preliminaries}

\subsection{Problem Formulation}
\label{sec:formulation}

\begin{table}
  \centering
  \caption{Notation list used in this paper.}
  \label{tab:notation}
  \setlength{\tabcolsep}{0.5\tabcolsep}
  \NiceMatrixOptions{notes/style=\arabic{#1}}
  \resizebox{\columnwidth}{!}{%
    \begin{NiceTabularX}{1.13\columnwidth}{cX}
      \toprule
      \textbf{Symbol} & \textbf{Description} \\ \midrule
      $N$ & Natural language question \\
      $Q$ & SQL query equivalent to question $N$ \\
      $\mathcal{D}$ & Set of databases available, each denoted by $D$ \\
      $T$ & Set of tables involved in query $Q$, with their properties \\
      $S$ & SQL query schema $\langle D, T \rangle$, where $T$ is the subset of tables in database $D$ \\
      $\mathcal{T}$ & Union of sets of tables from each database in $\mathcal{D}$ \\
      $\mathcal{S}$ & Schema space $\mathcal{D}\times\mathcal{P}(\mathcal{T})$\tabularnote{The cartesian product of $\mathcal{D}$ and $\mathcal{P}(\mathcal{T})$ is the set of all ordered pairs $\langle D, T \rangle$ that make up the solution (or decoding) space \textit{without any constraints}.}, where $\mathcal{P}(\mathcal{T})$ is the power set of $\mathcal{T}$ \\
      $\mathcal{G}$ & Schema graph $\langle \mathcal{V}, \mathcal{E} \rangle$, where $\mathcal{V}, \mathcal{E}$ are graph nodes and edges \\
      \bottomrule
    \end{NiceTabularX}%
  }
\end{table}

As mentioned above, we abstract and formulate the problem of natural language querying over massive databases and large numbers of tables with unspecified SQL query schema (\ie the target database and tables) as \emph{schema-agnostic \nlsql}.
Formally, as per the notation list detailed in \autoref{tab:notation}, the problem of \emph{schema-agnostic \nlsql} and its two subtasks \textit{schema routing} and \textit{SQL generation} are described in \autoref{def:text2sql}, \ref{def:schema-routing}, and \ref{def:sql-generation}, respectively.

\begin{definition}[\textsc{Schema-Agnostic \nlsql}]
  \label{def:text2sql}
  Given a natural language question $N$, schema-agnostic \nlsql generates an equivalent SQL query $Q$ that can be executed against the appropriate database $D$ drawn from a set of databases $\mathcal{D}$, without being provided with the intended SQL query schema $S=\langle D, T \rangle$.
\end{definition}

\begin{definition}[\textsc{Schema Routing}]
  \label{def:schema-routing}
  Schema routing identifies the SQL query schema $S=\langle D, T \rangle$ of a natural language question $N$ from the space of all possible schemata $\mathcal{S}=\mathcal{D}\times\mathcal{P}(\mathcal{T})$ within the specified set of databases $\mathcal{D}$ and tables $\mathcal{T}$.
\end{definition}

\begin{definition}[\textsc{SQL Generation}]
  \label{def:sql-generation}
  SQL generation is the process of generating a SQL query $Q$, given a natural language question $N$ and the intended SQL query schema $S=\langle D, T \rangle$.
\end{definition}

\stitle{Schema Linking vs. Schema Routing.}
Schema linking is an optional substep of \nlsql methods~\cite{DBLP:journals/vldb/KatsogiannisMeimarakisK23}.
Given an NL question and a database schema, schema linking aligns natural language terminologies, called ``mentions'', with their corresponding schema elements (such as tables, columns, and database values).
In this way, schema linking servers as a bridge between NL questions and database elements.

Schema routing is an extension of the schema linking concept adapted for querying over massive databases.
Specifically, schema routing differs from existing schema linking methods in the following aspects.
First, schema linking methods~\cite{DBLP:conf/emnlp/DongSLLZ19,DBLP:conf/emnlp/LeiWMGLKC20,DBLP:conf/aaai/Li00023} usually take the entire single-database schema and an NL question as input, while schema routing, driven by scalability requirements, should take the NL question as input with massive database schemata indexed.
Second, schema routing care about coarse-grained databases and tables, while leaving the fine-grained columns and values for further SQL generation.

\subsection{Generative Retrieval}

Traditional information retrieval (IR) techniques follow the ``index-retrieve'' paradigm.
Such a pipeline, which cannot be optimized together in an end-to-end manner, limits the final retrieval quality.
To address this concern, generative retrieval~\cite{DBLP:conf/nips/Tay00NBM000GSCM22,DBLP:conf/nips/WangHWMWCXCZL0022} (\aka DSI) is an emerging paradigm that leverages a Seq2Seq pre-trained language model (PLM) to directly generate relevant document IDs (\eg ``doc456'') in response to input queries, rather than vector representations for computing similarities.
Technically, DSI works by Seq2Seq learning on (query/doc\_chunk, doc\_id) pairs to model semantic mappings, completely parameterizing traditional ``index-retrieve'' pipelines within a single neural model.
Therefore, unlike conventional machine learning tasks that aim to generalize to unseen data, the training phase of DSI shifts to \emph{memorize} all retrieval targets -- including those used in evaluation -- similar to the indexing phase of traditional retrieval.
Critical ongoing research focuses on defining better document IDs~\cite{DBLP:conf/nips/0001YCWZRCYRR23,DBLP:journals/corr/abs-2305-13859}.
However, these studies mainly focus on retrieving elements individually without considering their relationships (\eg joinable relations between tables, triple relations in knowledge graphs, or semantic relations between documents).
To the best of our knowledge, this paper is the first to adapt this paradigm to jointly retrieve elements with relationships, opening up new opportunities for applying it in graph-based retrieval~\cite{DBLP:journals/corr/abs-2408-08921}.

\subsection{LLM-based Natural Language to SQL}

LLMs, developed by scaling up PLMs in terms of both model and data size, have emerged as a milestone in artificial intelligence~\cite{DBLP:journals/corr/abs-2303-18223}.
Despite their similar architecture and pre-training tasks, LLMs exhibit unexpected emergent abilities (\eg in-context reasoning, instruction following) compared to small PLMs~\cite{DBLP:journals/tmlr/WeiTBRZBYBZMCHVLDF22}.
These emergent abilities change the paradigm of model use from ``fine-tuning'' to ``prompt engineering'' that does not change the model parameters.
This shift has brought new opportunities for all kinds of natural language processing tasks, including \nlsql.
Coupled with well-designed schema-aware prompts and few-shot \nlsql examples, LLMs can generate SQL queries directly from NL questions, delivering state-of-the-art performance~\cite{DBLP:journals/pvldb/GaoWLSQDZ24,DBLP:journals/corr/abs-2306-08891,DBLP:journals/pacmmod/LiZLFZZWP0024}.
In addition, pipeline-based approaches decompose \nlsql into substeps including schema linking, query classification, query decomposition, SQL refinement, further extending the power of LLM-based \nlsql~\cite{DBLP:conf/nips/PourrezaR23,DBLP:journals/corr/abs-2403-09732}.
Nevertheless, LLM-based \nlsql still faces significant challenges in real-world scenarios, such as schema scalability, complex or ambiguous queries, SQL efficiency, and so on~\cite{DBLP:conf/cidr/FloratouPZDHTCA24,DBLP:conf/nips/LiHQYLLWQGHZ0LC23,DBLP:conf/emnlp/KothyariDSC23}.


\section{Methodology}

This section introduces \sys, a schema-agnostic \nlsql framework for effectively scaling natural language querying to massive databases.

\begin{figure}
  \centering
  \includegraphics[width=\linewidth]{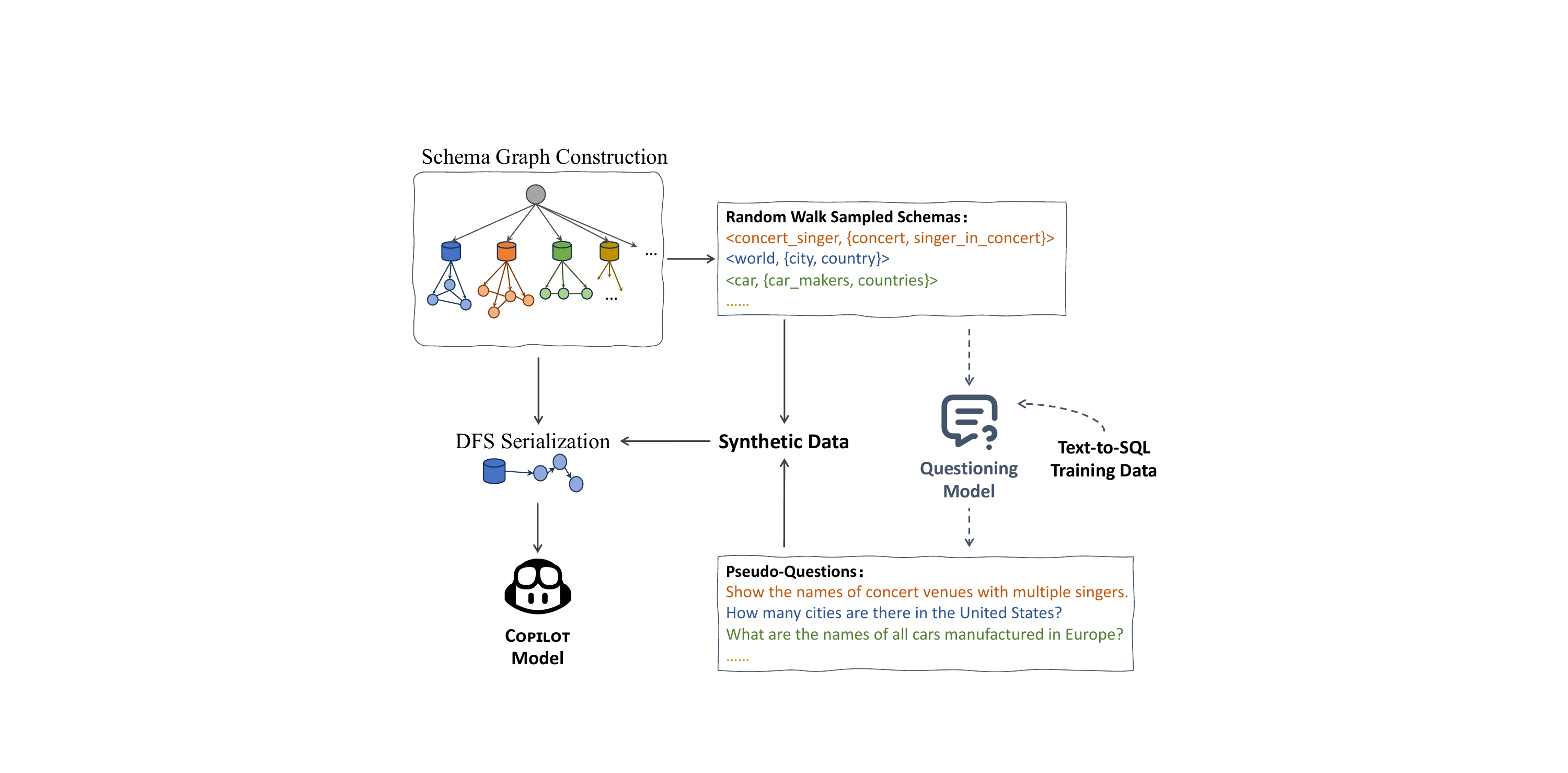}
  \caption{Training process of schema router. We first construct a schema graph to represent the relationships of all databases and their tables. Based on this graph, we sample SQL query schemata and generate pseduo-questions using a reverse-trained schema questioning model that synthesizes training data. Finally, we apply DFS serialization and train the schema router.}
  \label{fig:training}
\end{figure}

\subsection{Overview}

As shown in \autoref{fig:introduction}, \sys decouples natural language querying over massive databases into schema routing and SQL generation.
Specifically, given an NL question $N$, \sys utilizes a copilot model as a schema router to identify the SQL query schema $S=\langle D, T \rangle$ associated with that question.
After that, \sys utilizes an LLM to translate the NL question $N$ with the routed schema $S$ into a corresponding SQL query $Q$.
\autoref{ex:framework} provides an illustration of this process.
Consequently, this LLM-Copilot collaboration architecture enables easy integration with advanced LLM-based NL2SQL solutions and cost-effective scaling to massive databases for NLIDBs.

\begin{example}
  \label{ex:framework}
  Given a natural language question ``\emph{Which singers held concerts in 2022?}'', \sys converts it into an equivalent SQL query through the following two phases:

  \begin{enumerate}[nosep,wide,label=\arabic*.]
  \item \textsc{Schema Routing} identifies the relevant database and tables containing necessary information to answer the question.
\begin{lstlisting}[escapechar=\%]
  %\Schema{concert\_singer}{singer, singer\_in\_concert, concert}%
\end{lstlisting}

  \item \textsc{SQL Generation} prompts LLMs with the identified schema and NL question to generate the SQL query.
    \begin{lstlisting}[language=SQL]
  SELECT s.NAME
  FROM concert_singer.singer_in_concert AS sic
  JOIN concert_singer.singer AS s
    ON sic.singer_id = s.singer_id
  JOIN concert_singer.concert AS c
    ON sic.concert_id = c.concert_id
  WHERE c.year = 2022
    \end{lstlisting}
  \end{enumerate}
\end{example}

To better scale schema routing to massive databases by modeling relationships between schema elements and learning semantic mappings from natural language to schemata, we propose a relation-aware joint retrieval approach for end-to-end schema routing.
Specifically, we first construct a schema graph to represent the underlying relationships between databases and tables (\autoref{sec:schema-graph}).
Based on this graph, we design a schema serialization algorithm to serialize a SQL query schema into a token sequence so that it can be generated by Seq2Seq DSI (\autoref{sec:schema-serialization}).
To bridge the semantic gap between natural language and domain-specific schemata, we further propose a reverse generation paradigm for training data synthesis (\autoref{sec:data-synthesis}).
These measures allow us to effectively train the schema router with serialized pseudo-instances <$N$, $S$> and apply inference with graph-based constrained decoding (\autoref{sec:router-training-inference}).
\autoref{fig:training} illustrates the training process.
Finally, we explore various prompt strategies to select and incorporate multiple candidate schemata for LLM-based SQL generation (\autoref{sec:sql-generation}).

\subsection{Schema Graph Construction}
\label{sec:schema-graph}

Unlike natural language expressions, a valid SQL query must follow the logical data model~\cite{wiki:1152291307} of the database.
Therefore, it is necessary to inform the schema router about the inherent relationships and constraints between schema elements.
In general, the SQL query schema $S=\langle D, T \rangle$ involved in a valid SQL query should satisfy the following conditions.
First, the set of tables $T$ should be the subset of tables belonging to database $D$, since a SQL query should not reference tables that do not exist in the target database.
Second, the tables in $T$ should cross-reference at least one other table with a common column (\eg foreign key), so that the SQL query can connect them together via join or subquery clauses.
\autoref{ex:schema-graph-1} and \ref{ex:schema-graph-2} show two examples.
If two tables reference different columns of the junction table, then this junction table is required in a SQL query for these two tables.
However, if two tables reference the same column of another table, they can be linked implicitly.

\begin{example}
  \label{ex:schema-graph-1}
  Given a natural language question ``\emph{What are the names of the singers who performed in a concert in 2014?}'', the corresponding SQL query is:
  \begin{lstlisting}[language=SQL]
  SELECT s.name
  FROM singer_in_concert AS sc
  JOIN singer AS s ON sc.singer_id = s.singer_id
  JOIN concert AS c ON sc.concert_id = c.concert_id
  WHERE c.year = 2014
  \end{lstlisting}
  This query involves the \textsl{singer}, \textsl{concert}, and \textsl{singer\_in\_concert} tables, where \textsl{singer\_in\_concert} is the join table with two different columns referring to \textsl{singer} and \textsl{concert} respectively.
\end{example}

\begin{example}
  \label{ex:schema-graph-2}
  Given a natural language question ``\emph{Which rivers run through the state with the largest city in the US?}'', the corresponding SQL query is:
  \begin{lstlisting}[language=SQL]
  SELECT river_name FROM river
  WHERE traverse IN (
      SELECT state_name FROM city
      WHERE population = (
          SELECT MAX(population) FROM city
      )
  )
  \end{lstlisting}
  This query involves only the \textsl{city} and \textsl{river} tables, both of which refer to the primary key of the \textsl{state} table.
\end{example}

To this end, we construct a directed graph $\mathcal{G}=\langle \mathcal{V}, \mathcal{E} \rangle$ to represent the underlying schemata of all databases $\mathcal{D}$ and their tables $\mathcal{T}$.
Concretely, the schema graph consists of three types of nodes $\mathcal{V} = \{\nu_s\} \cup \mathcal{D} \cup \mathcal{T}$ and many types of edges $\mathcal{E}=\{e_{1}, \ldots, e_{\vert\mathcal{E}\vert}\}$, where $\nu_s$ is a special node denoting the \emph{set} of all databases $\mathcal{D}$.
Each edge $e \in \mathcal{E}$ in this graph represents a relation between two nodes that can be divided into two categories:

\begin{itemize}
\item \emph{Inclusion relations}: Inclusion relations indicate whether a database is part of the database collection and whether a table belongs to a database.
\item \emph{Table relations}: Table relations include explicit \textsc{Primary-Foreign}, implicit \textsc{Foreign-Foreign}, and \textsc{Joinable} relations between table nodes. \textsc{Joinable}~\cite{DBLP:conf/sigmod/SarmaFGHLWXY12} indicates that two tables share some values in certain attributes, allowing them to be combined together in SQL queries.
\end{itemize}

\begin{algorithm}[!t]
  \caption{Schema Graph Construction.}
  \label{alg:schema-graph}
  \DontPrintSemicolon
  \SetKwIF{If}{ElseIf}{Else}{if}{:}{elif}{else:}{}
  \SetKw{KwTo}{in}
  \SetKwFor{For}{for}{\string:}{}
  \SetKwFor{While}{while}{:}{}%

  \KwIn{A dictionary of all database schemata $databases$.}
  \KwOut{The heterogeneous directed schema graph $\mathcal{G}$.}
  \BlankLine
  $\mathcal{G} \leftarrow $ an empty directed graph \;
  \For{\textup{$database$ \KwTo $databases$}}{
    $links \leftarrow $ a dictionary with empty set as default value \;
    $\mathcal{G}$.addEdge($\nu_s$, $database$) \;
    \For{\textup{$table$ \KwTo $database$}} {
      $\mathcal{G}$.addEdge($database$, $table$) \;
      \tcp{\textsc{Joinable} covers the cases of \textsc{Primary-Foreign} and \textsc{Foreign-Foreign}.}
      \For{\textup{$relatedTable$ \KwTo getJoinableTables($table$)}} {
        $\mathcal{G}$.addEdge($table$, $relatedTable$)) \;
        $\mathcal{G}$.addEdge($relatedTable$, $table$))
      }
    }
  }
  \Return{$\mathcal{G}$} \;
\end{algorithm}

\autoref{alg:schema-graph} shows the procedure for constructing this three-tiered hierarchical graph.
By capturing the logical structure of database schemata in graph $\mathcal{G}$, any valid single-database SQL query schema $S$ is a \emph{trail}~\cite{wiki:1167548844} on this graph.
Each schema trail starts with node $\nu_s$, followed by a database node and some of its interconnected table nodes.
In this way, the schema graph facilitates relation-based serialization, valid SQL query schema sampling, and graph-based constrained decoding introduced below, serving as the foundation of \sys.

\subsection{Schema Serialization}
\label{sec:schema-serialization}

In this paper, we model schema routing as a generative retrieval process from an NL question $N$ to a SQL query schema $\langle D, T \rangle$.
Since a Seq2Seq LM takes a sequence of tokens as input and outputs a sequence of tokens, we need to design a schema serialization algorithm that converts the partially ordered SQL query schema into a sequence of tokens.

\stitle{Basic Serialization.}
A trivial solution is to randomly order the set of tables as DSI for ad-hoc retrieval~\cite{DBLP:conf/acl/ChenLH0S23} and identify the target database based on the votes of the generated tables after inference.
However, this serialization loses sight of the inclusion and table relations between schema elements.
Due to the position bias and autoregressive nature of Seq2Seq LMs, the order of schema elements is important for training routers to better model their relationships~\cite{DBLP:journals/tacl/LiuLHPBPL24,DBLP:conf/iclr/BerglundTKBSKE24}.
Therefore, we need a carefully designed serialization strategy that preserves relationships within SQL query schema.

\stitle{Depth-First Search Serialization.}
To address the above problem, we follow the constructed schema graph and design a depth-first search (DFS) serialization algorithm.
As shown in \autoref{alg:serialization}, we perform a DFS starting at node $\nu_s$ until all nodes of a SQL query schema have been visited and record the access order.
We then concatenate the sequence of node names into a sequence of tokens.
Formally, the \emph{depth-first search serialization} can be defined as:
\begin{equation*}
  \operatorname{Serialize}_\pi(S) \coloneqq \operatorname{Concat}\left( \left[\operatorname{DFS_\pi}(S, \mathcal{G})[i]\right]_{i=1}^{\left\vert\operatorname{DFS_\pi}(S, \mathcal{G})\right\vert} \right)
\end{equation*}
where $\pi$ is the node iteration order and $\operatorname{DFS_\pi}(S, \mathcal{G})$ is the DFS traversal sequence of schema $S$ on schema graph $\mathcal{G}$ when iterating successor nodes under order $\pi$ and $[i]$ indicates the $i$-th node of the traversal sequence.

We introduce the iteration order parameter $\pi$ to determine the DFS sequence because a SQL query schema is not a completely ordered set and there may be multiple DFS sequences for a SQL query schema.
We randomly select one DFS sequence at a time as the model target during training, and the schema router assigns different probabilities to different sequences originating from the same schema during inference.

\begin{algorithm}[!t]
  \caption{Depth-First Search Serialization.}
  \label{alg:serialization}
  \DontPrintSemicolon
  \SetKwIF{If}{ElseIf}{Else}{if}{:}{elif}{else:}{}
  \SetKw{KwTo}{in}
  \SetKwFor{For}{for}{\string:}{}
  \SetKwFor{While}{while}{:}{}%

  \KwIn{An ordered pair of a SQL query schema $S=\langle D, T \rangle$, the schema graph $\mathcal{G}$, and a node iteration order $\pi$.}
  \KwOut{The serialization of schema $S$.}
  \BlankLine
  $nodes$ $\leftarrow$ $\{ \nu_s \} \cup \{ D \} \cup T $ \;
  $visited$ $\leftarrow$ $[\;]$ \tcc*{an empty list}
  $stack$ $\leftarrow$ $[\nu_s]$ \;
  \While{\textup{$stack$ is non-empty}}{
    $node$ $\leftarrow$ $stack$.pop() \;
    $visited$.append($node$) \;
    \If{\textup{set($visited$) is equal to $nodes$}}{
      break \tcc*{cease the loop}
    }
    $successors$ $\leftarrow$ $[\;]$ \;
    \tcp{Iterates over successor nodes according to node iteration order $\pi$}
    \For{\textup{$successor$ \KwTo $\mathcal{G}$[$node$]}}{
      \If{\textup{$successor$ in $nodes$ and not in $visited$}}{
        $successors$.append($successor$) \;
      }
    }
    $stack$.extend($successors$) \;
  }
  \Return{$\operatorname{Concat}(visited[1: \operatorname{len}(visited)])$} \tcc*{skip $\nu_s$}
\end{algorithm}

\subsection{Training Data Synthesis}
\label{sec:data-synthesis}

Based on the schema graph and DFS serialization, we can train the schema router on question-schema pairs ($N$, $S$) as described in \autoref{sec:router-training-inference}.
However, it is infeasible to manually label training data for massive database schemata.
Furthermore, the generative retrieval paradigm prevents the schema router from generalizing to unseen schemata because they are not injected into model parameters.
To address these challenges, we introduce a labor-free training data synthesis method that automatically generates pseudo-instances via a reverse schema-to-question generation paradigm, as illustrated in \autoref{fig:training}.

Specifically, we sample a large number of valid SQL query schemata from the schema graph and generate a pseudo-question for each sampled schema.
These schemata are sampled by performing finite-length random walks starting at node $\nu_s$ over the schema graph, and the database and tables traversed by a walk form a sampled schema.
To generate an NL question for each sampled schema, we train a schema questioning model $\mathcal{M}_{\text{q}}$ in reverse based on the training sets of \nlsql datasets.
The reason for training a specific questioning model rather than using LLMs directly is to save the cost of synthesizing large amounts of training data.
We use a SQL parser to extract the metadata (tables and columns) of SQL queries from the training pairs ($N$, $Q$).
Unlike the schema router, which takes a condensed SQL query schema as output, the schema questioner accepts more detailed schema (\eg table columns and comments) as input to generate rich and detailed questions.
Given an illustration of the model input and output in \autoref{fig:s2q}, the training loss of the schema questioning model $\mathcal{M}_{q}$ with parameters $\theta_{\text{q}}$ on $m$ extracted schema-question pairs ($S$, $N$) is:
\begin{equation*}
  \mathcal{L}_{q} = - \sum_{i=1}^{m} \log P \left( N_i \mid S_i ; \theta_{\text{q}}\right)
\end{equation*}
In practice, it can also generate high-quality seed data from schemata through LLMs and further boost them in this way.

\begin{figure}
  \centering
  \resizebox{\columnwidth}{!}{%
    \small
    \begin{promptbox}{Pseudo-Question Generation}
      \textbf{Input:}
      Ask a question that needs to be answered by combining the contents of all the tables, based on the \texttt{word\textsubscript{1}} database schema provided below.

      \texttt{\ \ - countrylanguage(language, countrycode, official)}

      \hspace{9pt}\texttt{\ \ [Optional schema comments, if available.]}

      \texttt{\ \ - country(code, lifeexpectancy)}
      \tcblower
      \textbf{Output:} \textit{What is the official language of the country with the highest life expectancy?}
    \end{promptbox}
  }
  \caption{An illustration of pseudo-question generation.}
  \label{fig:s2q}
\end{figure}

\subsection{Schema Router Training \& Constrained Inference}
\label{sec:router-training-inference}

Through the above measures, we can efficiently train the schema router to parameterize the semantic mapping of massive database schemata and effectively route NL questions to their target schemata via graph-based constrained decoding.
Formally, the Seq2Seq schema router $\mathcal{M}_r$ takes a token sequence of question $N$ as input and generates the serialized schema sequence $\bar{S}$.
Suppose the synthetic data is $\{ (N_i, S_i) \mid 1 \le i \le n \}$, the training loss of schema router $\mathcal{M}_r$ with parameters $\theta_r$ is:
\begin{equation*}
  \mathcal{L}_r = - \sum_i^n \sum_\pi \log P \left( \bar{S}_i \mid N_i; \theta_r \right)
\end{equation*}

\begin{figure}
  \centering
  \includegraphics[width=\columnwidth]{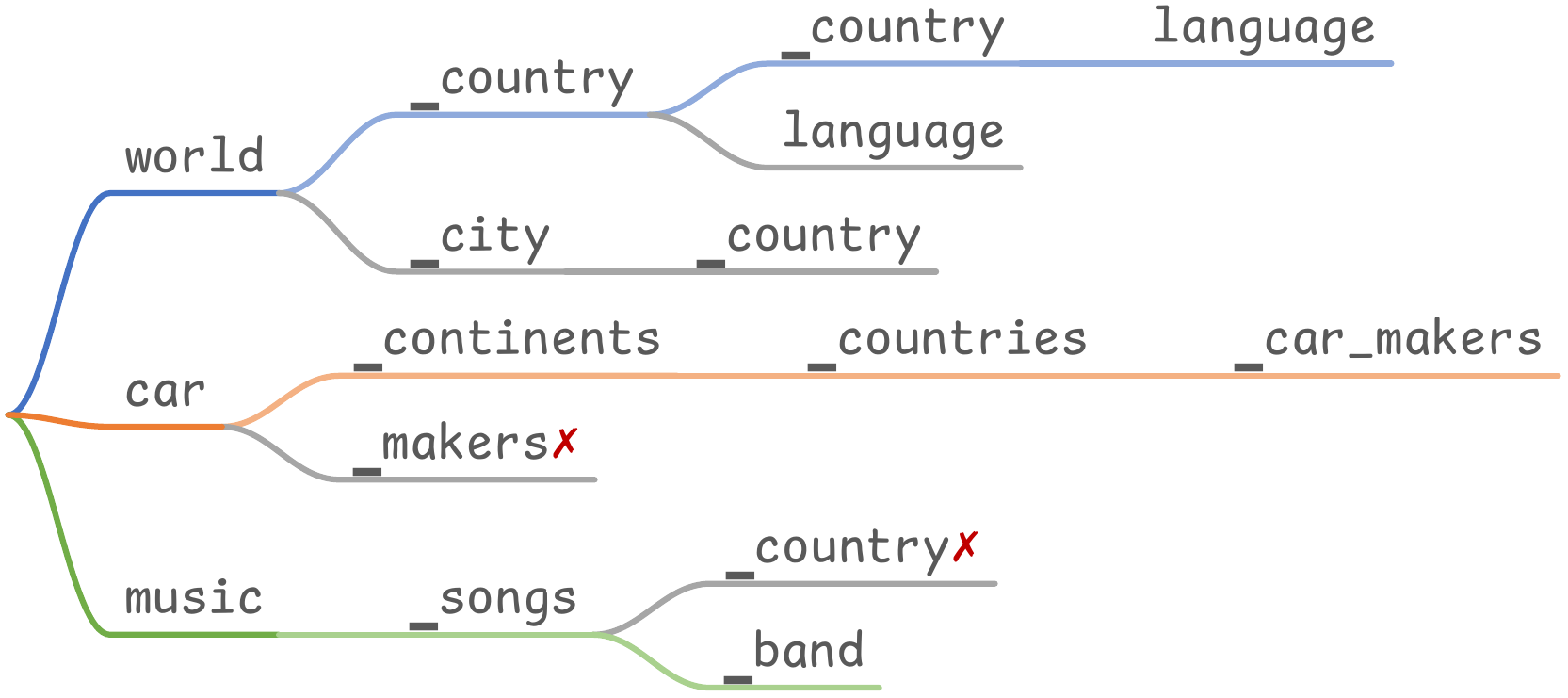}
  \caption{An example of graph-based constrained decoding. The underline (\textbf{\_}) denotes the separator between schema elements, and a red cross indicates that the generation of the token is invalid and disallowed. We leverage diverse beam search to ensure a more varied set of candidate sequences, represented in different colors in this graph.}
  \label{fig:decoding}
\end{figure}

After training the schema router, we further design a \emph{graph-based constrained decoding} algorithm to guarantee the validity of generated schemata.
As shown in \autoref{fig:decoding}, at each step of autoregressive decoding, we maintain a dynamic prefix tree containing the names of accessible nodes from decoded schema elements.
Unlike the DFS serialization, the accessible table nodes can only be the neighbors of generated tables, not the neighbors of the database, unless no tables have been generated yet.
Therefore, the available tokens (\ie part of the node name) for each decoding step can be obtained by searching the prefix tree, prefixed with the tokens generated after the last element separator.
Besides, multiple candidate schema sequences can be generated to improve the coverage of the target schema.
We thus utilize \emph{diverse beam search}~\cite{DBLP:journals/corr/VijayakumarCSSL16} to generate more diverse sets of candidate sequences and combine tables from schema sequences that share the same database to form the final candidate schemata.

\subsection{SQL Generation}
\label{sec:sql-generation}

\begin{figure}
  \centering
  \resizebox{\columnwidth}{!}{%
  \small
  \begin{promptbox}{Basic Prompt}
    \#\#\# Complete sqlite SQL query only and with no explanation \\
    \#\#\# Sqlite SQL tables, with their properties: \\
    \# \\
    \# \texttt{country(code, name, continent, region, ...)} \\
    \# \texttt{countrylanguage(countrycode, language, ...)} \\
    \# \\
    \#\#\# \textit{Which language is the most popular on the Asian continent?} \\
    \texttt{SELECT}
  \end{promptbox}
}
  \caption{An example of a basic \nlsql prompt.}
  \label{fig:basic-prompt}
\end{figure}

By decoupling schema-agnostic \nlsql into schema routing and SQL generation, \sys can easily integrate with orthogonal, advanced LLM-based \nlsql solutions, no matter whether in-context prompt engineering methods~\cite{DBLP:conf/nips/PourrezaR23,DBLP:journals/pvldb/GaoWLSQDZ24,DBLP:conf/pricai/GuoTTWWYW23,DBLP:conf/emnlp/ChangF23} or specific \nlsql LLMs~\cite{sqlcoder,nsql,DBLP:journals/pacmmod/LiZLFZZWP0024}.
However, since the schema router can generate multiple schemata from different databases for an NL question, we explore three prompt strategies to select and incorporate these candidate schemata for LLM-based \nlsql:

\begin{itemize}
\item \textbf{Best Schema Prompting}: This strategy instructs LLMs with the highest probability schema generated by the schema router, as shown in \autoref{fig:basic-prompt}. This basic prompt template has been shown to be the most effective prompt for SQL generation with OpenAI LLMs~\cite{DBLP:journals/pvldb/GaoWLSQDZ24}.
\item \textbf{Multiple Schema Prompting}: This strategy instructs LLMs with multiple schemata generated by the schema router for more possible coverage of the target schema. The prompt format is the same as the best schema prompting strategy and we concatenate multiple table schemata in the prompt.
\item \textbf{Multiple Schema COT Prompting}: This strategy instructs LLMs using multiple candidate schemata through chain-of-thought (COT) reasoning. As shown in \autoref{fig:cot-prompt}, the LLM is first asked to select the most relevant candidate schema and then generates SQL queries.
\end{itemize}

\begin{figure}
  \centering
  \resizebox{\columnwidth}{!}{%
  \small
  \begin{promptbox}{Chain-of-Thought Prompt}
    \underline{\normalsize Turn 1}:

    Based on the provided natural language question, find the database that can best answer this question from the list schemata below. Only output the corresponding database schema identifier in the [id] format, without any additional information.

    Question: \textit{Which language is the most popular on the Asian continent?}

    Sqlite SQL databases, with their tables and properties:
    \vspace{2pt}

    [1] \texttt{car} \\
    \texttt{continents(contid, continent)} \\
    \texttt{countries(countryid, countryname, continent)} \\
    \texttt{car\_makers(id, marker, fullname, country)}
    \vspace{1pt}

    [2] \texttt{word} \\
    \texttt{country(code, name, continent, region, ...)} \\
    \texttt{countrylanguage(countrycode, language, ...)}
    \vspace{1pt}

    [3] \texttt{...}
    [4] \texttt{...}
    [5] \texttt{...}

    \tcblower
    \small
    \underline{\normalsize Turn 2}: Filled basic prompt with \underline{Turn 1} selected schema.
  \end{promptbox}
}
  \caption{An example of a chain-of-thought \nlsql prompt with multiple candidate schemata.}
  \label{fig:cot-prompt}
\end{figure}


\section{Experiments}
\label{sec:experiments}

In this section, we conduct experiments to verify the feasibility and effectiveness of \sys.
First, we compare \sys with various retrieval-based methods for schema routing.
Then, we evaluate the performance of end-to-end schema-agnostic \nlsql.
Finally, we perform detailed ablation studies.

\subsection{Experimental Setup}

\subsubsection{Research Questions}

We conducted experiments to answer the following key research questions (RQs):

{\bf RQ1}: How does \sys compare with various retrieval baselines in schema routing?

{\bf RQ2}: Is \sys more robust against semantic mismatch between NL questions and schemata?

{\bf RQ3}: Does improving schema routing lead to better end-to-end schema-agnostic \nlsql?

{\bf RQ4}: What is the best prompt strategy for LLM-based SQL generation when there are multiple candidate schemata?

{\bf RQ5}: How does each module (\ie DFS serialization, data synthesis, decoding strategies) contribute to \sys?

\subsubsection{Datasets}
We adapt existing \nlsql datasets to evaluate schema routing and schema-agnostic \nlsql.

\begin{itemize}
\item \textbf{Spider}~\cite{DBLP:conf/emnlp/YuZYYWLMLYRZR18}: Spider is a widely-used, cross-domain \nlsql dataset with 10,181 questions and 5,693 unique SQL queries across 138 domains and 200 databases, requiring to generalize across different SQL queries and database schemas.
\item \textbf{Bird}~\cite{DBLP:conf/nips/LiHQYLLWQGHZ0LC23}: Bird is a challenging \nlsql dataset designed to emphasize the importance of database content, spotlighting the challenges presented by dirty database values, external knowledge requirements, and SQL efficiency.
\item \textbf{Fiben}~\cite{DBLP:journals/pvldb/SenLQOEDSMSS20}: Fiben is a dataset designed to simulate real-world data mart by IBM that closely mirrors actual enterprise data warehouses. Its schema conforms to standard financial ontologies, representing the complexity of schema routing in production environments.
  Fiben consists of 300 NL questions, corresponding to 237 complex SQL queries.
  These queries focus on business intelligence scenarios involving nesting and aggregation, making it an ideal testbed for evaluating \nlsql systems in real-world scenarios.
\item \textbf{Spider\textsubscript{syn}}~\cite{DBLP:conf/acl/GanCHPWXH20}: Spider\textsubscript{syn} is a dataset designed to evaluate the robustness of \nlsql models to synonym substitution, built on top of Spider with NL questions modified by replacing schema-related words with real-world paraphrases.
\item \textbf{Spider\textsubscript{real}}~\cite{DBLP:conf/naacl/DengAMPSR21}: Spider\textsubscript{real} is another robustness dataset created from the dev split of the Spider dataset by manually modifying the original NL questions to remove the explicit mention of column names. 
\end{itemize}

\begin{table}
  \centering
  \caption{Statistics of datasets.}
  \label{tab:datasets}
  \setlength{\tabcolsep}{0.8\tabcolsep}
  \resizebox{\columnwidth}{!}{%
    \begin{NiceTabular}{ccccccc}
      \toprule
      \multirow{2}{*}{\textbf{Dataset}} & \multicolumn{3}{c}{\textbf{Size}} & \multicolumn{3}{c}{\textbf{Schema}}  \\
      \cmidrule(lr){2-4} \cmidrule(lr){5-7}
      & \textbf{Train} & \textbf{Dev.} & \textbf{Test} & \textbf{\# DBs} & \textbf{\# Tables}  & \textbf{\# Cols} \\
      \cmidrule(lr){1-7}
      Spider                     & 7000           & --            & 1034\tabularnote{The original development (dev.) sets are used as test sets.}         & 166   & 876        & 4503        \\
      Bird                       & 9427           & --            & 1534\tabularnote{The original development (dev.) sets are used as test sets.}         &  80   & 597        & 4337        \\
      Fiben                      & --              & --            & 279           &   1   & 152        & 374         \\
      Spider\textsubscript{syn}  & 7000           & --            & 1034          & 166   & 876        & 4503        \\
      Spider\textsubscript{real} & --              & --            & 508           & 166   & 876        & 4503        \\
      \bottomrule
    \end{NiceTabular}%
  }
\end{table}

\begin{table*}
  \centering
  \caption{Schema routing performance on regular test sets.}
  \label{tab:regular-test}
  \begin{NiceTabular}[tabularnote = {$^{\natural}$ and $^{\sharp}$ indicate statistically significant improvements over the best LLM-enhanced sparse retrieval baseline (\ie CRUSH\textsubscript{BM25}) at $t$-test $p < 0.05$ and $p < 0.01$, respectively. $^\dagger$ and $^\ddagger$ indicate statistically significant improvements over the best dense and fine-tuned retrieval baseline (\ie DTR) at $t$-test $p < 0.05$ and $p < 0.01$, respectively. The best results are in \textbf{bold} and the second best results are {\ul underlined}. Database recall of BM25 is less than 100\% for Fiben due to vocabulary mismatch between some NL questions and the entire database schema, leading to retrieval failures.}]{ccccccccccccc}
    \toprule
    \multirow{4}{*}{\textbf{Method}} & \multicolumn{4}{c}{\textbf{Spider}}                                        & \multicolumn{4}{c}{\textbf{Bird}}                                          & \multicolumn{4}{c}{\textbf{Fiben}}                                         \\
    \cmidrule(lr){2-5} \cmidrule(lr){6-9} \cmidrule(lr){10-13}
    & \multicolumn{2}{c}{\textbf{Database}} & \multicolumn{2}{c}{\textbf{Table}} & \multicolumn{2}{c}{\textbf{Database}} & \multicolumn{2}{c}{\textbf{Table}} & \multicolumn{2}{c}{\textbf{Database}} & \multicolumn{2}{c}{\textbf{Table}} \\
    \cmidrule(lr){2-3} \cmidrule(lr){4-5} \cmidrule(lr){6-7} \cmidrule(lr){8-9} \cmidrule(lr){10-11} \cmidrule(lr){12-13}
    & \textbf{R@1}      & \textbf{R@5}      & \textbf{R@5}     & \textbf{R@15}   & \textbf{R@1}      & \textbf{R@5}      & \textbf{R@5}     & \textbf{R@15}   & \textbf{R@1}      & \textbf{R@5}      & \textbf{R@5}     & \textbf{R@15}   \\
    \midrule
    \multicolumn{13}{c}{\textit{Zero-shot}} \\
    BM25                             & 70.12             & 91.49             & 86.49            & 93.87           & 57.11             & 90.16             & 68.32            & 82.81           & 98.92             & 98.92             & 33.34            & 38.62           \\
    SXFMR                            & 62.57             & 89.46             & 80.42            & 92.39           & 66.49             & 88.20             & 67.56            & 83.05           & 100.0             & 100.0             & 28.21            & 46.47           \\
    \cmidrule{1-13}
    \multicolumn{13}{c}{\textit{LLM-enhanced}} \\
    CRUSH\textsubscript{BM25}        & {\ul 72.34}       & {\ul 94.49}       & {\ul 87.19}      & {\ul 95.06}     & 56.58             & 94.85             & 68.37            & 87.82           & 100.0             & 100.0             & 34.87            & {\ul 54.03}     \\
    CRUSH\textsubscript{SXFMR}       & 63.25             & 91.97             & 82.21            & 93.86           & 70.14             & 92.44             & 70.58            & 85.07           & 100.0             & 100.0             & 34.06            & 50.78           \\
    \cmidrule{1-13}
    \multicolumn{13}{c}{\textit{Fine-tuned}} \\
    BM25                             & 68.28             & 91.78             & 86.60            & 93.87           & 60.63             & 90.48             & 69.87            & 83.34           & 98.92             & 98.92             & 32.67            & 38.30           \\
    DTR                              & 61.51             & 92.84             & 76.27            & 93.18           & {\ul 76.99}       & \textbf{97.33}    & {\ul 76.24}      & {\ul 91.96}     & 100.0             & 100.0             & {\ul 37.72}      & 48.85           \\
    \sys                             & \textbf{85.01}$^{\sharp\ddagger}$    & \textbf{96.42}$^{\ddagger}$    & \textbf{91.63}$^{\ddagger}$   & \textbf{97.51}$^{\natural\ddagger}$  & \textbf{88.92}$^{\sharp\ddagger}$    & {\ul 97.13}$^{\sharp}$      & \textbf{85.83}$^{\sharp\ddagger}$   & \textbf{94.59}$^{\sharp}$  & \textbf{100.0}    & \textbf{100.0}    & \textbf{41.12}$^{\sharp\dagger}$   & \textbf{56.89}$^{\ddagger}$  \\
    \bottomrule
  \end{NiceTabular}%
\end{table*}


\stitle{Dataset Adaptation.}
These datasets are designed to evaluate \nlsql on a single database, where each question is annotated with its target database.
To approach the schema-agnostic \nlsql, we make several adjustments.
First, we remove the single-database constraint and treat all databases in the collection as potential query targets.
Next, we extract metadata (tables and columns) from each SQL query using a SQL parser~\cite{sqlglot} and exclude any queries that cannot be parsed.
Finally, we combine the target database and extracted metadata as a SQL query schema, along with the original NL question and SQL query, to form each instance ($N$, $S$, $Q$).
This results in three database collections (Spider, Bird, and Fiben), with instances divided into training and test sets as in the original setup.
The robustness variants of Spider share the same database collection.
\autoref{tab:datasets} summarizes the dataset statistics after adaptation.

\subsubsection{Baselines}

We compare \sys to various sparse and dense retrieval techniques for schema routing (RQ1 \& RQ2), including zero-shot, LLM-enhanced, and fine-tuned methods:

\begin{itemize}
\item \textbf{BM25}~\cite{wiki:1180365145}: BM25 is a renowned ranking function that serves as a standard in search engines to rank documents by their relevance to a specified query.
\item \textbf{SXFMR}~\cite{DBLP:conf/emnlp/ReimersG19}: SXFMR is a technique for obtaining generic embedding models for dense retrieval by contrastive learning of Transformer(XFMR)-based PLMs on related text pairs.
\item \textbf{CRUSH}~\cite{DBLP:conf/emnlp/KothyariDSC23}: CRUSH adopts a two-stage approach to retrieve schema elements for \nlsql. It first instructs an LLM to generate a hallucinated schema, and then combines multiple retrieval results and reranks them based on their relationships to obtain the actual schema.
\item \textbf{DTR}~\cite{DBLP:conf/naacl/HerzigMKE21}: DTR is an approach that uses contrastive learning on (question, table) pairs to develop table retrievers for open-domain question answering over tables.
\end{itemize}

For schema-agnostic \nlsql (RQ3 \& RQ4), since existing \nlsql methods cannot be directly applied to this problem even when equipped with schema linking (as discussed in \autoref{sec:formulation}), we combine a representative LLM-based \nlsql method~\cite{DBLP:journals/pvldb/GaoWLSQDZ24} with different retrieval methods for comparison. This allows us to isolate and evaluate the effects of schema routing and prompt strategies on end-to-end SQL generation in a controlled setting.

\subsubsection{Evaluation Metrics}

Different evaluation metrics are used for schema routing and SQL generation respectively.

For \emph{schema routing}, we use Recall@\textit{k} and mAP (mean average precision) to evaluate the relevance of retrieved databases and tables.
Recall@\textit{k} measures the fraction of relevant instances in the top-\textit{k} retrieved candidates.
mAP calculates the average precision over all queries, which reflects the overall retrieval performance.

For \emph{SQL generation}, traditional surface-level metrics are inappropriate for LLM-based approaches as they may produce SQL queries that differ greatly from the ground truth.
Following recent studies~\cite{DBLP:journals/corr/abs-2306-08891,DBLP:journals/pvldb/GaoWLSQDZ24,DBLP:conf/nips/LiHQYLLWQGHZ0LC23}, we use EX (execution accuracy), which compares the results of generated SQL queries with corresponding ground truths.
We also report the cost (\$) of LLM invocations.

\subsubsection{Implementation Details}

We implemented \sys using Python 3.10, Pytorch 2.1, along with some major libraries such as Transformers~\cite{DBLP:conf/emnlp/WolfDSCDMCRLFDS20} and Pytorch Lightning~\cite{Falcon_PyTorch_Lightning_2019}. The source code is available at \href{https://github.com/tshu-w/DBCopilot}{github.com/tshu-w/DBCopilot}.

\stitle{Schema Routing.}
We adopted T5-base~\cite{DBLP:journals/jmlr/RaffelSRLNMZLL20} as the backbone of the schema questioner and router.
We trained a unified schema questioner on Spider and Bird \nlsql training sets and three schema routers on each database collection (Spider, Bird, and Fiben).
For each database collection, we sampled $1 \times 10^5$ SQL query schemata uniformly from the constructed graph, covering all (100\%) databases and tables, and synthesized the corresponding training instances using the schema questioner.
We set the batch size to $32$, the learning rate to $5 \times 10^{-5}$, and the maximum number of training epochs to $20$ for all models.
Model parameters were optimized using AdamW~\cite{DBLP:conf/iclr/LoshchilovH19} and a linear learning rate schedule with no warm-up steps.
To detect joinable tables besides primary and foreign, we adopted a heuristic way that two tables are joinable if the exact match overlap (Jaccard similarity) of their column values is greater than 0.85.
During schema routing, we generated 10 schema sequences for each question using diverse beam search, with beams and beam groups set to 10 and the diversity penalty set to 2.0.

\stitle{SQL Generation.}
We used gpt-3.5-turbo-0125\footnote{\url{https://platform.openai.com/docs/models\#gpt-3-5-turbo}} to select the optimal schema and generated the SQL query, with the generation temperature set to 0 for reproducibility.

\stitle{Baselines.}
To maintain fairness, all baselines compared in this study relied on schema information only.
We treated tables as retrieval targets, with the content of each table being the flat normalized names of the table and its columns.
For each NL question, we retrieved the top 100 tables and ranked the databases based on the average score of their retrieved tables.
We used the standard \emph{Okapi BM25} with two adjustable parameters for BM25 and adopted the best performing \emph{all-mpnet-base-v2} for SXFMR.
BM25 and DTR were fine-tuned on synthetic data consistent with \sys to verify our relation-aware joint retrieval.

All experiments were run using Docker image nvidia/cuda:12.1.1-cudnn8-devel-ubuntu22.04 on a server with AMD EPYC 7H12 64-Core Processors, 1TB RAM, and NVIDIA A100 40G GPUs.

\subsection{Schema Routing (RQ1 \& RQ2)}

\subsubsection{Experimental Results}

\begin{figure}
  \centering
  \includegraphics[width=\columnwidth]{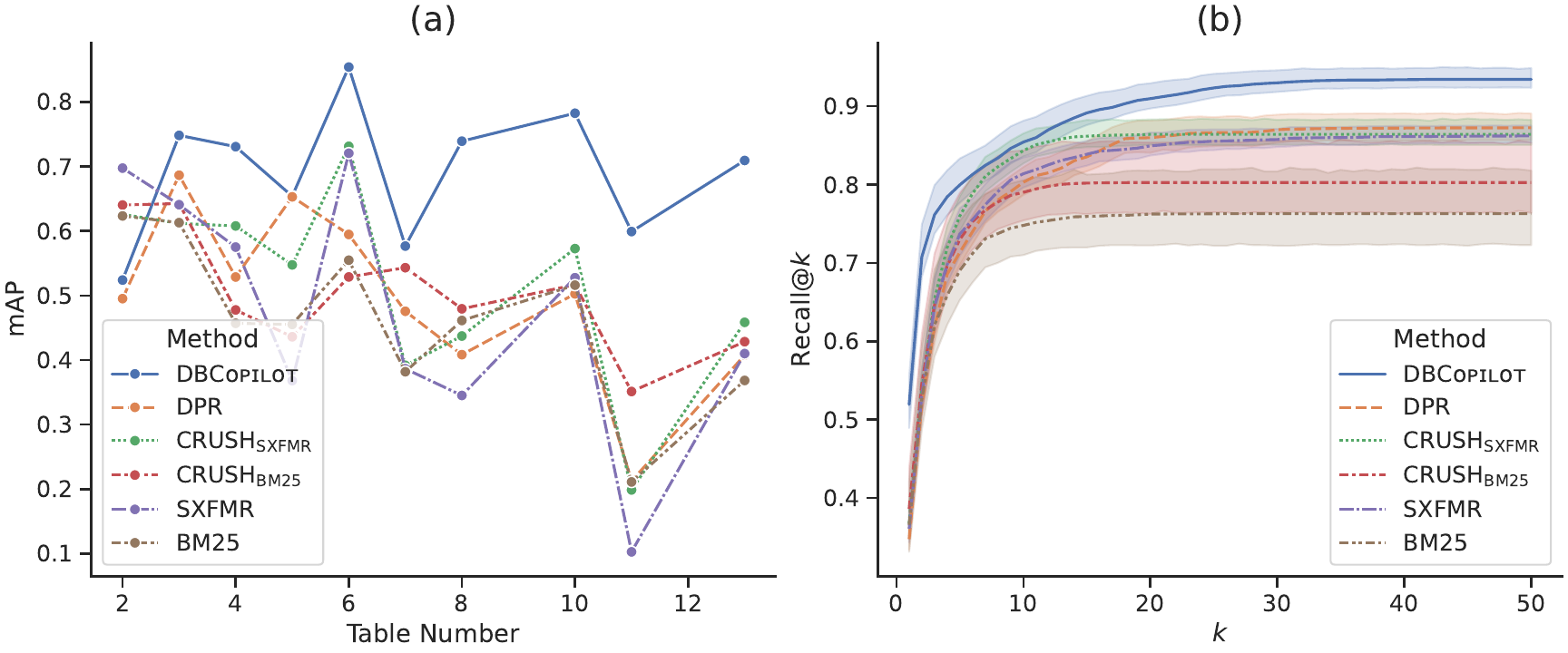}
  \caption{(a) Table mAP \wrt different numbers of database tables. (b) Table recall@\textit{k} \wrt different numbers of retrieved tables.}
  \label{fig:table-number}
\end{figure}

We first compare the performance of schema routing on both regular and robustness datasets, with the experimental results shown in \autoref{tab:regular-test}, \autoref{tab:robutness-test}, \autoref{fig:table-number}, and the following findings.

\autoref{tab:regular-test} shows the database and table recall on regular test sets.
We can see that \sys consistently outperforms all zero-shot, LLM-enhanced, or fine-tuned baselines by a large margin.
The database recall@1 of \sys surpasses the second-best approach by up to 12.67\% and 11.93\% on the Spider and Bird datasets, respectively.
This result verifies that \sys can effectively identify the target database of an NL question.
Furthermore, \sys also holds an advantage in table recall, with a lead of up to 9.59\%, which is crucial for providing complete tables for SQL generation.
Even on the single database dataset Fiben, \sys still achieves a 3.4\% improvement.
These results confirm that \sys can effectively navigate NL questions to their target databases and tables over massive databases.

\noindent{\textbf{Finding 1. \sys consistently outperforms all types of baselines by a large margin in schema routing.}}

When querying over massive databases, data consumers without schema knowledge typically express NL questions that are highly inconsistent with database schemata.
To evaluate the robustness of different methods in this situation, \autoref{tab:robutness-test} shows the experimental results on Spider\textsubscript{syn} and Spider\textsubscript{real}, which replace explicitly mentioned schema elements in NL questions posed by \nlsql annotation.
We can see that the performance of all methods decreases in this actual scenario setting.
Sparse retrieval baselines are significantly affected as they rely heavily on the vocabulary matching, and the LLM-enhanced method also struggles with this challenge, despite being designed specifically for this case.
Compared to these baselines, \sys is the least affected and therefore maintains a larger advantage with 14.89\% and 19.88\% on database recall@1, as well as 2.83\% and 6.94\% on table recall@5.
We believe that by injecting schema knowledge through schema-specific training, \sys can better understand the semantics of database schemata and effectively learn the semantic mapping between NL questions and schemata.

\begin{table}
  \centering
  \caption{Schema routing performance on robustness tests.}
  \label{tab:robutness-test}
  \resizebox{\columnwidth}{!}{%
    \setlength{\tabcolsep}{0.2\tabcolsep}
    \begin{NiceTabular}[tabularnote = {We use the same indicators and markups as \autoref{tab:regular-test}.}]{@{}ccccccccc@{}}
      \toprule
      \multirow{4}{*}{\textbf{Method}} & \multicolumn{4}{c}{\textbf{Spider\textsubscript{syn}}}                     & \multicolumn{4}{c}{\textbf{Spider\textsubscript{real}}}                    \\
      \cmidrule(lr){2-5} \cmidrule(lr){6-9}
      & \multicolumn{2}{c}{\textbf{Database}} & \multicolumn{2}{c}{\textbf{Table}} & \multicolumn{2}{c}{\textbf{Database}} & \multicolumn{2}{c}{\textbf{Table}} \\
      \cmidrule(lr){2-3} \cmidrule(lr){4-5} \cmidrule(lr){6-7} \cmidrule(lr){8-9}
      & \textbf{R@1}      & \textbf{R@5}      & \textbf{R@5}     & \textbf{R@15}   & \textbf{R@1}      & \textbf{R@5}      & \textbf{R@5}     & \textbf{R@15}   \\
      \midrule
      \multicolumn{9}{c}{\textit{Zero-shot}} \\
      BM25                             & 29.11             & 58.03             & 45.30            & 58.87           & 52.95             & 81.1              & 70.87            & 80.64           \\
      SXFMR                            & {\ul 47.78}       & 80.85             & 65.90            & 82.96           & 58.86             & 88.78             & 79.48            & 90.52           \\
      \cmidrule{1-9}
      \multicolumn{9}{c}{\textit{LLM-enhanced}} \\
      CRUSH\textsubscript{BM25}        & 35.88             & 65.18             & 46.80            & 64.06           & 58.46             & 86.42             & 75.38            & 89.14           \\
      CRUSH\textsubscript{SXFMR}       & 45.45             & 81.14             & {\ul 67.52}      & {\ul 84.18}     & {\ul 59.65}       & {\ul 91.14}       & {\ul 80.99}      & {\ul 92.78}     \\
      \cmidrule{1-9}
      \multicolumn{9}{c}{\textit{Fine-tuned}} \\
      BM25                             & 30.66             & 57.83             & 46.12            & 59.65           & 50.00             & 81.30             & 70.77            & 80.87           \\
      DTR                              & 46.52             & {\ul 82.69}       & 65.09            & 83.19           & 59.65             & 89.96             & 73.23            & 91.73           \\
      \sys                             & \textbf{62.67}$^{\sharp\ddagger}$    & \textbf{85.11}$^{\sharp}$    & \textbf{70.35}$^{\sharp\ddagger}$   & \textbf{86.26}$^{\sharp\ddagger}$  & \textbf{79.53}$^{\sharp\ddagger}$    & \textbf{95.47}$^{\sharp\ddagger}$    & \textbf{87.93}$^{\sharp\ddagger}$   & \textbf{96.06}$^{\sharp\ddagger}$  \\
      \bottomrule
    \end{NiceTabular}%
  }
\end{table}


\noindent{\textbf{Finding 2. \sys is more robust when NL questions are highly inconsistent with database schemata.}}

We further analyzed the performance regarding different database sizes (\ie the number of tables contained) and the number of tables retrieved.
Experimental results are shown in \autoref{fig:table-number} (a) and (b), respectively.
First, we categorize the instances based on the number of tables contained in their target database and compare the table mAP across different database sizes.
As shown in the \autoref{fig:table-number} (a), the mAP of \sys is relatively stable as the number of database tables increases, while the independent retrieval baselines decrease significantly.
We believe this is because by modeling schema routing as a relation-aware joint retrieval process, \sys can better capture the relationships within the database compared to independent retrieval methods, which is critical for routing databases that contain many tables.
Furthermore, as shown in \autoref{fig:table-number} (b), the table recall of \sys is consistently higher than that of the baselines as the number of retrieved tables increases, achieving a recall@20 that exceeds the recall@50 of all baselines.
This suggests that retrieving more tables and reranking them based on their relationships can hardly compensate for the advantages of our end-to-end relation-aware joint retrieval approach.

\noindent{\textbf{Finding 3. Relation-aware joint retrieval are essential for DBCopilot to perform effective schema routing.}}

\subsubsection{Case Study}
\label{sec:case-study}

\begin{figure}
  \centering
  \small
  \begin{casebox}[colback=green!3!white]
    \textbf{Question:}
    Which language is the most popular on the Asian continent?

    \textbf{SQL Query:}
    \begin{lstlisting}[language=SQL,basicstyle=\linespread{0.8}\selectfont]
  SELECT T2.Language
  FROM country AS T1 JOIN countrylanguage AS T2
    ON T1.Code = T2.CountryCode
  WHERE T1.Continent = "Asia"
  GROUP BY T2.Language
  ORDER BY COUNT(*) DESC
  LIMIT 1
    \end{lstlisting}

    \textbf{SQL Schema:} $\:\:\:\:\:$\Schema{word}{country, countrylanguage}

    \tcblower

    \begin{tabular}{@{}ll@{}}
      \textbf{BM25} & \Schema{car}{continents, countries} \\
      \textbf{SXFMR} & \Schema{roller\_coaster}{country} \\
      \textbf{CRUSH\textsubscript{BM25}} & \Schema{car}{continents, countries} \\
      \textbf{CRUSH\textsubscript{SXFMR}} & \Schema{match\_season}{country} \\
      \textbf{DTR} & \Schema{world}{countrylanguage} \\
      \textbf{\sys} & \Schema{world}{countrylanguage, country} \\
    \end{tabular}
  \end{casebox}
  \caption{Successful case of \sys in schema routing.}
  \label{fig:successful-case}
\end{figure}

We provide successful cases to illustrate how \sys improves schema routing through schema-specific training and joint retrieval.
As shown in \autoref{fig:successful-case}, \sys is the only approach that accurately routes the appropriate schema for the given question.
Sparse retrieval methods (\ie BM25 and CRUSH\textsubscript{BM25}) rely heavily on lexical matching while ignoring the semantic matching between the question and the schemata.
However, untuned dense retrieval methods (\ie SXFMR and CRUSH\textsubscript{SXFMR}) also fail to capture the underlying semantics of tables and retrieve the wrong databases.
Fine-tuning based approaches identify the correct database where DTR misses the \texttt{country} table linked to the \texttt{country\-language} table, but our relation-aware \sys does not.

While \sys consistently outperforms the baselines, there are a few cases where DTR retrieves the correct schema but \sys does not.
For example, \autoref{fig:failure-case} shows that \sys confuses the \texttt{airlines} and \texttt{airports} tables of database \texttt{flight}, but DTR covers the correct tables by retrieving more loosely.
Upon further analysis, we find that the quality of synthetic data impacts the robustness of schema routing.
This is attributed to two main factors: 1) hallucination~\cite{DBLP:conf/nips/PapicchioPC23}, where the generated NL questions deviate from sampled schemata; 2) generation bias, where the simple pipeline lacks diversity and quality control mechanisms.
Developing more sophisticated data synthesis techniques to address these challenges remains an important direction for future work.

\begin{figure}
  \centering
  \small
  \begin{casebox}[colback=red!3!white]
    \textbf{Question:}
    How many flights fly from Aberdeen to Ashley?

    \textbf{SQL Query:}
    \begin{lstlisting}[language=SQL]
  SELECT COUNT(*) FROM FLIGHTS AS T1
  JOIN AIRPORTS AS T2
    ON T1.DestAirport = T2.AirportCode
  JOIN AIRPORTS AS T3
    ON T1.SourceAirport = T3.AirportCode
  WHERE T2.City = "Ashley" AND T3.City = "Aberdeen"
    \end{lstlisting}

    \textbf{SQL Schema:} $\:\:\:$\Schema{flight}{airports, flights}

    \tcblower

    \begin{tabular}{@{}ll@{}}
      \textbf{DTR}  & \Schema{flight}{flights, airports, airlines} \\
      \textbf{\sys} & \Schema{flight}{flights, airlines} \\
    \end{tabular}
  \end{casebox}
  \caption{Failure case where DTR retrieves the correct tables but \sys fails. This is because the schema questioning model incorrectly generates similar but wrong questions for the \Schema{flight\textsubscript{2}}{flights, airlines} schema.}
  \label{fig:failure-case}
\end{figure}

\begin{table}
  \centering
  \caption{Method efficiency \& resource consumption.}
  \label{tab:resource-consumption}
  \resizebox{\columnwidth}{!}{%
    \setlength{\tabcolsep}{0.5\tabcolsep}
    \NiceMatrixOptions{notes/style=\arabic{#1}}
    \begin{NiceTabular}{@{}cccccc@{}}
      \toprule
      \textbf{Method} & \textbf{QPS}\tabularnote{Queries per second (QPS) and memory consumption is calculated with a query batch size of 64.} & \textbf{Build (s)}\tabularnote{Build time includes the time for model training and index construction.}  & \textbf{Disk (MB)} & \textbf{RAM (MB)} & \textbf{GPU (MB)} \\
      \midrule
      BM25            & 1677.1       & 10                  & 36                 & 800               & -                 \\
      SXFMR           & 49.0         & 48                  & 420                & 2,230             & 3742              \\
      \cmidrule(lr){1-6}
      CRUSH\textsubscript{BM25}\tabularnote{Computations are performed on the same device except for CRUSH, which requires commercial LLM inference.}
                      & 0.9          & 10                  & 36                 & 1104              & -                 \\
      CRUSH\textsubscript{SXFMR}\tabularnote{Computations are performed on the same device except for CRUSH, which requires commercial LLM inference.}
                      & 0.8          & 48                  & 420                & 2502              & 3742              \\
      \cmidrule(lr){1-6}
      BM25            & 1677.1       & 6972                & 36                 & 800               & -                 \\
      DTR             & 49.0         & 5161                & 420                & 2,230             & 3742              \\
      \sys            & 43.0         & 13678               & 852                & 656               & 11807             \\
      \bottomrule
    \end{NiceTabular}%
  }
\end{table}

\subsubsection{Efficiency \& Resource Consumption}

We also measure the efficiency and resource consumption of each schema routing method.
\autoref{tab:resource-consumption} shows the number of queries per second (QPS), the duration of build time, the amount of disk space used, and the consumption of system and GPU memory.
LLM inference and complex post-processing are the main reasons for the slowness of CRUSH approaches.
\sys, based on an encoder-decoder architecture, consumes more disk space and GPU memory than dense retrieval approaches that based on an encoder-only architecture.
Despite the slightly higher resource consumption for training and inference in generative retrieval, we believe and have observed that the growing popularity of generative AI is leading to an increasing number of proposed optimizations.

\subsection{Schema-Agnostic \nlsql (RQ3 \& RQ4)}

This section compares the performance of schema-agnostic \nlsql under various situations, including providing gold schemata, using different schema routing methods, and prompt strategies. Experimental results are shown in \autoref{tab:sql-generation}.

\etitle{Oracle Test.}
We first conduct oracle tests to access the upper bound of schema-agnostic \nlsql, as well as the impact of irrelevant schema elements for LLM-based \nlsql, by providing gold schemata at various small scales.
Specifically, we gradually move from 5 database schemata to the gold database, then to gold tables only, or even to gold tables with only gold columns.
Experimental results in \autoref{tab:sql-generation} show that as the number of schema elements increases, the EX of LLM-generated SQL queries continues to decrease significantly, while the inference cost keeps increasing many times over.
This suggests that LLMs, like PLMs~\cite{DBLP:conf/aaai/Li00023}, are also affected by extraneous schema elements in SQL generation.
Predictably, the situation gets worse as the schema scales. 
As a result, it is highly inefficient and ineffective to expose massive database schemata directly to LLMs, making schema routing necessary for schema-agnostic \nlsql.

\begin{table}
  \centering
  \caption{Evaluation of schema-agnostic \nlsql with various gold schemata, different schema routing methods, and prompt strategies.}
  \label{tab:sql-generation}
  \resizebox{\columnwidth}{!}{%
    \setlength{\tabcolsep}{0.6\tabcolsep}
    \NiceMatrixOptions{notes/style=\arabic{#1}}
    \begin{NiceTabular}[tabularnote = {
        ``T.'' and ``C.'' stand for table and column, respectively.
        We bold best results of schema-agnostic \nlsql and underline the best with human in the loop to select the optimal candidate schema.
        For retrieval baselines, a candidate schema consists of the top candidate database and the retrieved tables that belong to the database.
      }]{@{}cccccccc@{}}
      \toprule
      \multirow{3}{*}{\textbf{\begin{tabular}[c]{@{}c@{}}Routing\\ Method\end{tabular}}} & \multirow{3}{*}{\textbf{\begin{tabular}[c]{@{}c@{}}Prompt\\ Template\end{tabular}}} & \multicolumn{2}{c}{\textbf{Spider}} & \multicolumn{2}{c}{\textbf{Bird}} & \multicolumn{2}{c}{\textbf{Spider\textsubscript{syn}}} \\
      \cmidrule(lr){3-4} \cmidrule(lr){5-6} \cmidrule(lr){7-8}
                                                                                         &                                                                                     & \textbf{EX}       & \textbf{Cost}   & \textbf{EX}      & \textbf{Cost}  & \textbf{EX}        & \textbf{Cost}     \\
      \midrule
      \multicolumn{8}{c}{\textit{Oracle Test}} \\
      \multicolumn{1}{l}{}                                                               & Gold T. \& C.                                                                                                                                             & 81.24             & 0.07            & 35.98            & 0.15           & 77.37              & 0.07              \\
      \multicolumn{1}{l}{}                                                               & Gold T.                                                                                                                                                        & 76.60             & 0.09            & 31.03            & 0.24           & 68.67              & 0.09              \\
      \multicolumn{1}{l}{}                                                               & Gold DB                                                                                                                                                      & 74.08             & 0.12            & 27.57            & 0.40           & 61.99              & 0.13              \\
      \multicolumn{1}{l}{}                                                               & 5 DB w. Gold                                                                                                                                                & 70.21             & 0.13            & 18.38            & 8.80           & 51.64              & 0.40              \\
      \midrule
      \multicolumn{8}{c}{\textit{Best Schema Prompting}} \\
      CRUSH\textsubscript{BM25}                                                          & Top 1                                                                                                                                                              & 56.58             & 0.10            & 14.73            & 0.26           & 20.60              & 0.09              \\
      DTR                                                                                & Top 1                                                                                                                                                              & 43.91             & 0.11            & 19.04            & 0.33           & 26.21              & 0.10              \\
      \sys                                                                               & Top 1                                                                                                                                                              & \textbf{63.35}    & 0.11            & \textbf{23.47}   & 0.36           & \textbf{37.43}     & 0.11              \\
      \cmidrule(lr){1-8}
      \multicolumn{8}{c}{\textit{Multiple Schema Prompting}} \\
      CRUSH\textsubscript{BM25}                                                          & Top 5                                                                                                                                                              & 55.61             & 0.20            & 17.60            & 0.57           & 21.86              & 0.19              \\
      DTR                                                                                & Top 5                                                                                                                                                              & 54.45             & 0.27            & 19.56            & 0.79           & 32.59              & 0.27              \\
      \sys                                                                               & Top 5                                                                                                                                                              & 61.51             & 0.31            & 21.45            & 0.95           & 35.49              & 0.31              \\
      \cmidrule(lr){1-8}
      \multicolumn{8}{c}{\textit{Multiple Schema COT Prompting}} \\
      CRUSH\textsubscript{BM25}                                                          & Top 5                                                                                                                                                              & 53.97             & 0.28            & 22.10            & 0.83           & 22.34              & 0.26              \\
      DTR                                                                                & Top 5                                                                                                                                                              & 50.68             & 0.36            & 22.43            & 1.08           & 28.53              & 0.37              \\
      \sys                                                                               & Top 5                                                                                                                                                              & 51.26             & 0.40            & 22.75            & 1.24           & 31.24              & 0.39              \\
      \cmidrule(lr){1-8}
      \multicolumn{8}{c}{\textit{Human in the Loop}} \\
      CRUSH\textsubscript{BM25}                                                          & Top 5                                                                                                                                                              & 69.34             & 0.10            & 25.10            & 0.32           & 34.72              & 0.09              \\
      DTR                                                                                & Top 5                                                                                                                                                              & 66.92             & 0.11            & 25.42            & 0.37           & 47.78              & 0.11              \\
      \sys                                                                               & Top 5                                                                                                                                                              & {\ul 72.05}       & 0.11            & {\ul 26.27}      & 0.37           & {\ul 55.42}        & 0.11              \\
      \bottomrule
    \end{NiceTabular}%
  }
\end{table}


\noindent{\textbf{Finding 4. Schema routing is necessary for scaling natural language querying to massive databases.}}

\etitle{Impact of Improved Recall.}
We then perform schema-agnostic \nlsql to verify the impact of improved recall on the accuracy of generated SQL queries.
Without loss of generality, we compare \sys to sparse CRUSH\textsubscript{BM25} and dense DTR baselines, which achieve suboptimal routing performance on Spider and Bird, respectively.
As shown in \autoref{tab:sql-generation}, \sys outperforms CRUSH\textsubscript{BM25} and DTR by 4.43\%\textasciitilde11.22\% EX when using the best schema prompting strategy and by 0.85\%\textasciitilde20.07\% EX when employing human in the loop to select the optimal schema from candidates.
We compare and discuss the prompting strategies in the next paragraph.
The LLM inference cost of \sys is slightly higher, possibly due to its successful retrieval of more required tables.
The correlation between schema routing recall and SQL generation EX can also be confirmed by comparing these two baselines.
CRUSH\textsubscript{BM25} outperforms DTR in recall and EX on Spider, while DTR achieves higher EX on the other two datasets due to its superior recall.
Thus, the success of schema routing is essential for schema-agnostic \nlsql.

\noindent{\textbf{Finding 5. \sys leads in schema-agnostic \nlsql due to its superior schema routing performance.}}

\etitle{Exploration of Prompt Strategies.}
By routing multiple candidate schemata, the recall of the target schemata for NL questions can be greatly improved.
We explore three prompt strategies for leveraging multiple candidate schemata during SQL generation.
As shown in \autoref{tab:sql-generation}, when schema routing through \sys, the performance of \emph{multiple schema prompting} is slightly lower than that of \emph{single schema prompting}.
However, when schema routing through baselines, the performance of \emph{multiple schema prompting} is typically better than that of \emph{single schema prompting}.
This is due to the trade-off between the improved recall brought by multiple candidate schemata and the impact of extraneous schemata on LLMs.
As for \emph{multiple schema COT prompting}, this strategy shows no absolute advantage over \emph{multiple schema prompting}, and each wins and loses on different routing methods and datasets.
Thus, the limitations of LLMs in distinguishing domain-specific schemata make it cost effective for \sys to find the correct schema in the top 1 candidate.
Finally, by introducing human-in-the-loop to select the desired schema from top 5 candidates, \sys comes close to the ideal performance of providing the gold database schema, but with lower costs.

\noindent{\textbf{Finding 6. The best schema prompting strategy is optimal for \sys unless there is human interaction.}}

\subsection{Ablation Study (RQ5)}

\begin{table}
  \centering
  \caption{Performance loss in ablation studies.}
  \label{tab:ablation-study}
  \resizebox{\columnwidth}{!}{%
    \setlength{\tabcolsep}{0.5\tabcolsep}
    \NiceMatrixOptions{notes/style=\arabic{#1}}
    \begin{NiceTabular}{@{}lcccccccc@{}}
      \toprule
      & \multicolumn{4}{c}{\textbf{Spider}}                                        & \multicolumn{4}{c}{\textbf{Bird}}                                          \\
      \cmidrule(lr){2-5} \cmidrule(lr){6-9}
      & \multicolumn{2}{c}{\textbf{Database}} & \multicolumn{2}{c}{\textbf{Table}} & \multicolumn{2}{c}{\textbf{Database}} & \multicolumn{2}{c}{\textbf{Table}} \\
      \cmidrule(lr){2-3} \cmidrule(lr){4-5} \cmidrule(lr){6-7} \cmidrule(lr){8-9}
      & \textbf{R@1}      & \textbf{R@5}      & \textbf{R@5}    & \textbf{R@15}    & \textbf{R@1}      & \textbf{R@5}      & \textbf{R@5}    & \textbf{R@15}    \\
      \midrule
      \sys      & 85.01             & 96.42             & 91.63           & 97.51            & 88.92             & 97.13             & 85.83           & 94.59            \\
      \quad w/ BS\tabularnote{``BS'', ``OD'', ``MD'', ``CD'', and ``DB'' stand for basic serialization, original \nlsql training data, mixed synthetic and original training data, constrained decoding, and diverse beam search.}     & -3.20             & -2.90             & -4.09           & -5.75            & -3.39             & -1.63             & -6.43           & -5.99            \\
      \quad w/ OD\tabularnote{``BS'', ``OD'', ``MD'', ``CD'', and ``DB'' stand for basic serialization, original \nlsql training data, mixed synthetic and original training data, constrained decoding, and diverse beam search.}\tabularnote{Note that \sys trained on original \nlsql training data fails to work because training and test sets do not share the same databases, and generative retrieval is limited to generalizing to unseen schemata. However, according to baselines fine-tuned on synthetic data in \autoref{tab:regular-test} and \autoref{tab:robutness-test}, as well as ablation results on other modules, we would like to emphasize that data synthesis is not the only effective part.}     & -84.72            & -82.40            & -88.76          & -79.55           & -88.92            & -82.27            & -85.73          & -88.81           \\
      \quad w/ MD\tabularnote{``BS'', ``OD'', ``MD'', ``CD'', and ``DB'' stand for basic serialization, original \nlsql training data, mixed synthetic and original training data, constrained decoding, and diverse beam search.}     & -0.48             & +0.20             & -0.83           & -0.64            & -3.85             & -0.72             & -3.39           & -0.81            \\
      \quad w/o CD\tabularnote{``BS'', ``OD'', ``MD'', ``CD'', and ``DB'' stand for basic serialization, original \nlsql training data, mixed synthetic and original training data, constrained decoding, and diverse beam search.}    & -0.39             & -0.19             & -4.15           & -2.51            & +0.12             & -0.45             & -2.81           & -1.93            \\
      \quad w/o DB\tabularnote{``BS'', ``OD'', ``MD'', ``CD'', and ``DB'' stand for basic serialization, original \nlsql training data, mixed synthetic and original training data, constrained decoding, and diverse beam search.}    & +0.48             & -3.00             & +0.64           & -2.88            & -0.52             & -4.89             & -3.69           & -8.87            \\
      \bottomrule
    \end{NiceTabular}%
  }
\end{table}


We perform ablation studies on \sys to validate the contribution of each module, as summarized in \autoref{tab:ablation-study}.

\etitle{DFS Serialization.} DFS serialization is essential for schema routers to learn the underlying relations of databases and tables.
To demonstrate this, we replace the DFS serialization with the unordered basic serialization of tables, similar to DSI for multiple documents.
As shown in \autoref{tab:ablation-study}, this results in a 1\% to 6\% reduction in recall.
In addition, DFS serialization facilitates graph-based constrained decoding to reduce the search space, either by first generating a database and then generating its tables, or by further generating based on existing tables.

\begin{figure}
  \centering
  \includegraphics[width=\columnwidth]{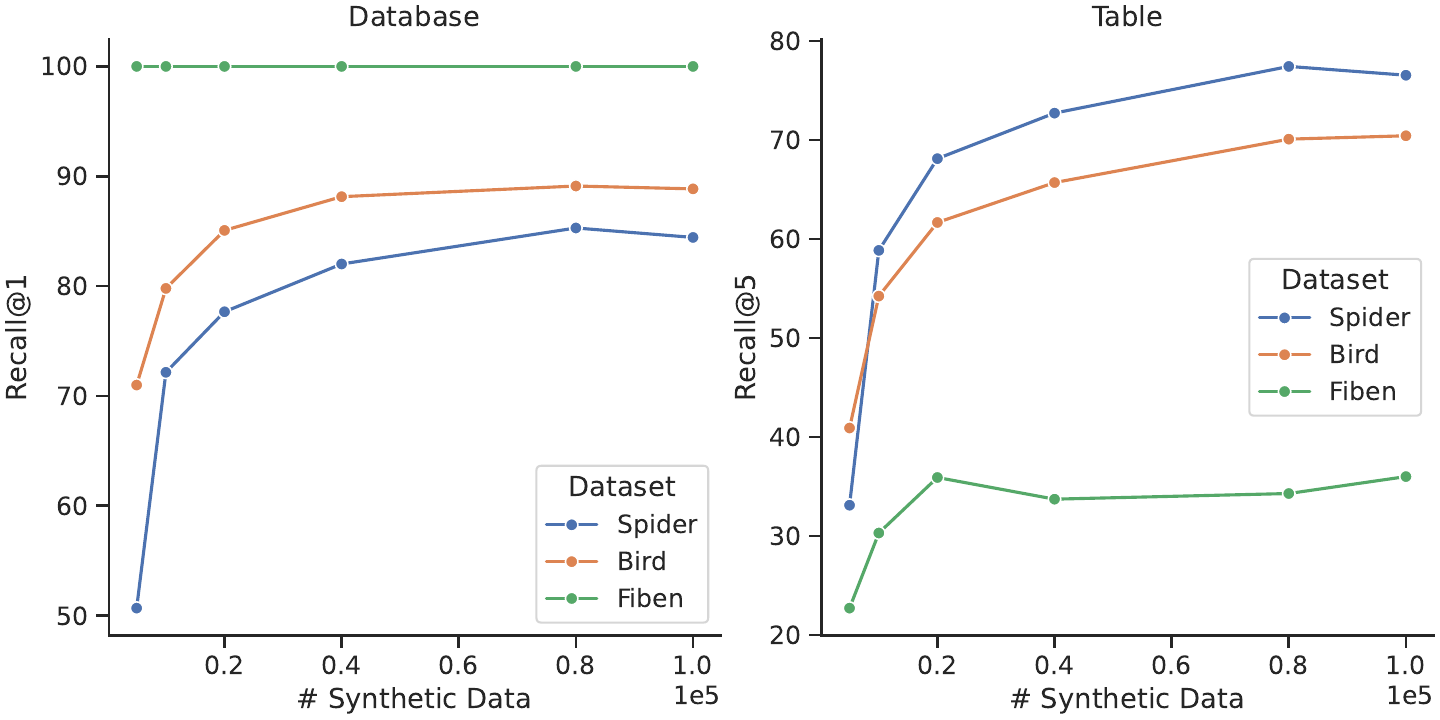}
  \caption{Database recall@1 and table recall@5 \wrt the number of (\#) synthetic training data.}
  \label{fig:ablation-study}
\end{figure}

\etitle{Data Synthesis.} Generative retrieval enables relation-aware end-to-end schema routing, but is unable to generalize to unseen schemata during training.
We can see that routers trained on instances ($N$, $S$) extracted from original \nlsql datasets are almost completely incapable of routing on unseen test schemata.
Data synthesis by sampling schemata and generating pseudo-questions can effectively inject semantic mappings to model parameters.
Besides, mixing original and synthetic data for training does not lead to more accurate schema routing, possibly due to label imbalance.
\autoref{fig:ablation-study} shows how the performance varies with the amount of synthetic data.
Database and table recalls increase rapidly at first as more data is synthesized, but level off after a certain amount.
How to improve the quality of synthetic data and reduce the need for its quantity may be a future direction.

\etitle{Decoding Strategy.} Decoding optimizations (\ie constrained decoding and group beam search) are beneficial for generative schema routing during inference.
Specifically, constrained decoding is useful for generating accurate table identifiers, since removing it mainly affects table recalls, about 2\% to 4\%.
Meanwhile, group beam search helps to generate multiple diverse schemata because database recall@5 decreases significantly without it.

\noindent{\textbf{Finding 7. DFS serialization, data synthesis, and decoding optimizations contribute in different ways.}}

\section{Related Work}

\subsection{Natural Language to SQL}
\nlsql is an essential part of natural language interfaces to databases, enabling non-experts to easily query structured data.
Most research has approached this task along the lines of semantic parsing~\cite{DBLP:conf/akbc/KamathD19,DBLP:conf/coling/LiQH20}.
Early \nlsql parsing systems adopted symbolic approaches that relied heavily on rule or grammar engineering, making them difficult to apply widely~\cite{DBLP:conf/aaai/ZelleM96,DBLP:conf/afips/Woods73}.
In recent years, with the advances in deep learning, especially PLMs, \nlsql approaches that using neural generation models have become the mainstream~\cite{DBLP:conf/aaai/Li00023,DBLP:conf/acl/WangSLPR20,DBLP:journals/pacmmod/GuF00JM023,DBLP:conf/aaai/LiHCQ0HHDSL23,DBLP:conf/iclr/0009WLWTYRSX21}.
Some efforts have been made to improve encoding and decoding, such as relation-aware encoding~\cite{DBLP:conf/acl/WangSLPR20,DBLP:conf/aaai/LiHCQ0HHDSL23} and grammar-based decoding~\cite{DBLP:journals/corr/abs-1711-04436,DBLP:conf/emnlp/ScholakSB21}.
Some other efforts have focused on generating accurate SQL queries through schema linking~\cite{DBLP:conf/aaai/Li00023,DBLP:conf/acl/WangSLPR20} or skeleton generation~\cite{DBLP:journals/pacmmod/GuF00JM023,DBLP:journals/corr/abs-2306-08891}.

LLMs have emerged as a new paradigm for \nlsql~\cite{DBLP:journals/corr/abs-2204-00498,DBLP:journals/corr/abs-2303-13547,DBLP:journals/pvldb/GaoWLSQDZ24,DBLP:journals/pvldb/LiLCLT24}.
Unlike previous work, a core component of the LLM-based \nlsql solution is prompt engineering, \ie prompting LLM to generate correct and efficient SQL queries.
Several efforts have been made to select appropriate in-context learning examples through question classification~\cite{DBLP:conf/nips/PourrezaR23}, skeleton retrieval~\cite{DBLP:conf/pricai/GuoTTWWYW23}, and in-domain example synthesis~\cite{DBLP:conf/emnlp/ChangF23}.
Research has also been done on decoupling zero-shot \nlsql into PLM-based sketch generation and LLM-based SQL completion~\cite{DBLP:journals/corr/abs-2306-08891}.
Recently, pipeline-based approaches decompose \nlsql into steps including schema linking, query classification, query decomposition, SQL refinement, further extending the power of LLM-based \nlsql~\cite{DBLP:conf/nips/PourrezaR23,DBLP:journals/corr/abs-2403-09732}.
These studies are orthogonal to our work and can be easily integrated to optimize SQL generation of \sys.

\subsection{Table Discovery}
Table discovery techniques are critical to data management, allowing users to explore and gain insights from data warehouses or lakes~\cite{DBLP:conf/sigmod/Fan00M23}.
A common practice for table discovery is query-driven discovery~\cite{DBLP:conf/sigmod/ZhangI20,DBLP:journals/pvldb/NargesianZMPA19}, where the query is typically in the form of keywords, tables, or NL questions.
For keyword search, data discovery systems find relevant tables by searching for topic terms in the metadata~\cite{DBLP:journals/pvldb/CafarellaHK09,DBLP:journals/pvldb/PimplikarS12}.
Table understanding~\cite{DBLP:journals/pvldb/DengSL0020}, including domain discovery~\cite{DBLP:conf/kdd/LiHG17,DBLP:journals/pvldb/OtaMFS20} and table annotation~\cite{DBLP:journals/pvldb/LimayeSC10,DBLP:journals/pvldb/VenetisHMPSWMW11}, has emerged as a means to mitigate the missing, incomplete, and inconsistent table schemata for keyword search.
For table search, tables are retrieved to augment a given table as the query~\cite{DBLP:conf/sigmod/SarmaFGHLWXY12}.
Joinable table search~\cite{DBLP:conf/icde/DongT0O21,DBLP:conf/sigmod/ZhuDNM19} and unionable table search~\cite{DBLP:journals/pvldb/FanWLZM23,DBLP:journals/pacmmod/KhatiwadaFSCGMR23} aim to augment a query table with additional attributes and new tuples, respectively.
For NL question search, open domain question answering retrieves independent tables to answer the NL question with table-based models~\cite{DBLP:conf/iclr/ChenCSWC21,DBLP:conf/naacl/HerzigMKE21}.
Although dense retrieval methods based on contrastive learning have achieved advanced performance in the above tasks~\cite{DBLP:conf/icde/DongT0O21,DBLP:journals/pvldb/FanWLZM23,DBLP:conf/naacl/HerzigMKE21}, they are difficult to consider the structural relationships between tables and retrieve them jointly.

Alternatively, data navigation allows users to navigate interactively through the tables organized hierarchically~\cite{DBLP:conf/sigmod/NargesianPZBM20} or relationally~\cite{DBLP:conf/icde/FernandezAKYMS18}.
Our work can be seen as automated data navigation, where we employ relation-aware DSI to model the query-based navigation in an end-to-end manner.
There is another line of research that focuses on table content discovery for Web table QA~\cite{DBLP:journals/tacl/Badaro0P23}, which typically employs PLMs to manipulate and analyze HTML tables of limited size (with average \#row < 20)~\cite{DBLP:journals/pacmmod/HulsebosDG23}. However, these systems are not suitable for NLIDBs as they cannot efficiently handle large relational tables with thousands or millions of rows and diverse data types that are better handled through semantic parsing.


\section{Discussion \& Future Work}

\sys has made significant progress in schema routing and schema-agnostic \nlsql. However, there are some limitations that need to be addressed in future work.

\begin{enumerate}[wide,label=\arabic*)]
\item \textbf{Cross-Databases Querying}.
  While current system shows the capability to route NL questions over massive databases, it is limited to generating SQL queries against a single database.
  Extending \sys to support cross-database SQL querying faces several challenges:
  i) the heterogeneity of different databases and the complexity of ETL processes for cross-database SQL querying;
  ii) the increased difficulty in discovering implicit relationships between tables across multiple databases than within a single database;
  iii) the absence of comprehensive benchmarks for evaluating cross-database \nlsql systems.
  We envision an alternative direction to decompose complex NL questions into multiple sub-questions, each answerable by a single database, and then integrate the results.
\item \textbf{Synthetic Data Quality}.
  As discussed in the case study, the quality of synthetic data used for training has a notable impact on router performance.
  Future efforts could focus on: i) generating high-quality synthetic data by using larger LLMs with better factual consistency; ii) providing more schema descriptions to reduce hallucination; iii) implementing automated validation mechanisms using TNLI~\cite{DBLP:journals/pacmmod/BussottiVSP23} to identify and remove low-quality synthetic data.
  In addition, reducing the amount of data used to train schema routers while maintaining their performance is also a worthwhile exploration.
\item \textbf{Dynamic Schema Update}.
  In real-world scenarios, database schemata may evolve over time, which creates a need for all retrieval-based methods to update the index.
  Although we have decoupled the dynamic schema routing into a compact and flexible copilot model, and it currently takes less than 4 hours to train a schema router from scratch on thousands of tables, managing to incremental update DSI can further reduce the computational resources required for evolving schemata.
\end{enumerate}

\section{Conclusion}

In this paper, we present \sys, a framework for effectively scaling natural language querying to massive databases.
\sys decouples schema-agnostic \nlsql into schema routing and SQL generation through \textsc{LLM-Copilot} collaboration.
We present an approach to effectively navigate the target database and tables for an NL question in a relation-aware, end-to-end manner.
We also propose a training data synthesis paradigm to adapt routers over massive databases without manual intervention, and explore various prompt strategies to incorporate multiple candidate schemata in LLM-based SQL generation.
Extensive experiments verify the effectiveness of our solution.
In summary, \sys pushes the boundaries of \nlsql, enabling researchers to better strategize their quest for data accessibility.


\begin{acks}
  We sincerely thank the reviewers for their insightful comments and valuable suggestions.
  This work was supported in part by Beijing Natural Science Foundation under Grant L243006, and in part by the Natural Science Foundation of China under Grant No. 62476265 and 62306303.
\end{acks}

\clearpage

\bibliographystyle{ACM-Reference-Format}
\bibliography{src/ref}



\end{document}